\pdfoutput=1

\documentclass[11pt]{article}

\usepackage[]{acl}

\usepackage{times}
\usepackage{latexsym}
\usepackage{booktabs}
\usepackage{multirow}
\usepackage{multicol}
\usepackage[T1]{fontenc}

\usepackage[utf8]{inputenc}
\usepackage{graphicx}

\usepackage{microtype}

\usepackage{algorithm}
\usepackage{algorithmicx}
\usepackage{amsmath}
\usepackage{algpseudocode}
\usepackage{amsfonts,amssymb}
\usepackage{svg}
\usepackage{hyperref}
\title{Hierarchical Verbalizer for Few-Shot Hierarchical Text Classification}

\author{
   Ke Ji\textsuperscript{1,2}\thanks{\; Work done during an internship at Xiaobing.AI},
   Yixin Lian\textsuperscript{2},
   Jingsheng Gao\textsuperscript{2},
   Baoyuan Wang\textsuperscript{2}\thanks{\; Corresponding Author}
   \\
    \textsuperscript{\rm 1} School of Computer Science and Engineering, Southeast University, China\\
 \textsuperscript{\rm 2} Xiaobing.AI \\
    {\tt keji@seu.edu.cn } \\
        {\tt \{lianyixin, gaojingsheng, wangbaoyuan\}@xiaobing.ai }
}

\newcommand{\setParDis}{\setlength {\parskip} {0.3cm} }
\newcommand{\setParDef}{\setlength {\parskip} {0pt} }

\begin{document}
\maketitle
\begin{abstract}
Due to the complex label hierarchy and intensive labeling cost in practice, the hierarchical text classification (HTC) suffers a poor performance especially when low-resource or few-shot settings are considered. 
Recently, there is a growing trend of applying prompts on pre-trained language models (PLMs), which has exhibited effectiveness in the few-shot flat text classification tasks. 
However, limited work has studied the paradigm of prompt-based learning in the HTC problem when the training data is extremely scarce. 
In this work, we define a path-based few-shot setting and establish a strict path-based evaluation metric to further explore few-shot HTC tasks. 
To address the issue, we propose the hierarchical verbalizer ("HierVerb"), a multi-verbalizer framework treating HTC as a single- or multi-label classification problem at multiple layers and learning vectors as verbalizers constrained by hierarchical structure and hierarchical contrastive learning. 
In this manner, HierVerb fuses label hierarchy knowledge into verbalizers and remarkably outperforms those who inject hierarchy through graph encoders, maximizing the benefits of PLMs. 
Extensive experiments on three popular HTC datasets under the few-shot settings demonstrate that prompt with HierVerb significantly boosts the HTC performance, meanwhile indicating an elegant way to bridge the gap between the large pre-trained model and downstream hierarchical classification tasks.
\footnote{Our code and few-shot dataset are publicly available at \href{https://github.com/1KE-JI/HierVerb}{https://github.com/1KE-JI/HierVerb}.}

\end{abstract}

\section{Introduction}


Hierarchical text classification (HTC) is a long-standing research problem due to the wide range of real applications \citep{mao-etal-2019-hierarchical}. 
However, prior works could still suffer poor performance in practice due to the nature of its sophisticated label hierarchy as well as the requirement of large-scale data annotation before training the model. 
Therefore, solving the HTC under the low-resource \citep{wang2022hpt} or few-shot setting becomes an urgent research topic.


\begin{figure}[!t]
\centering
\includegraphics[width=0.44\textwidth]{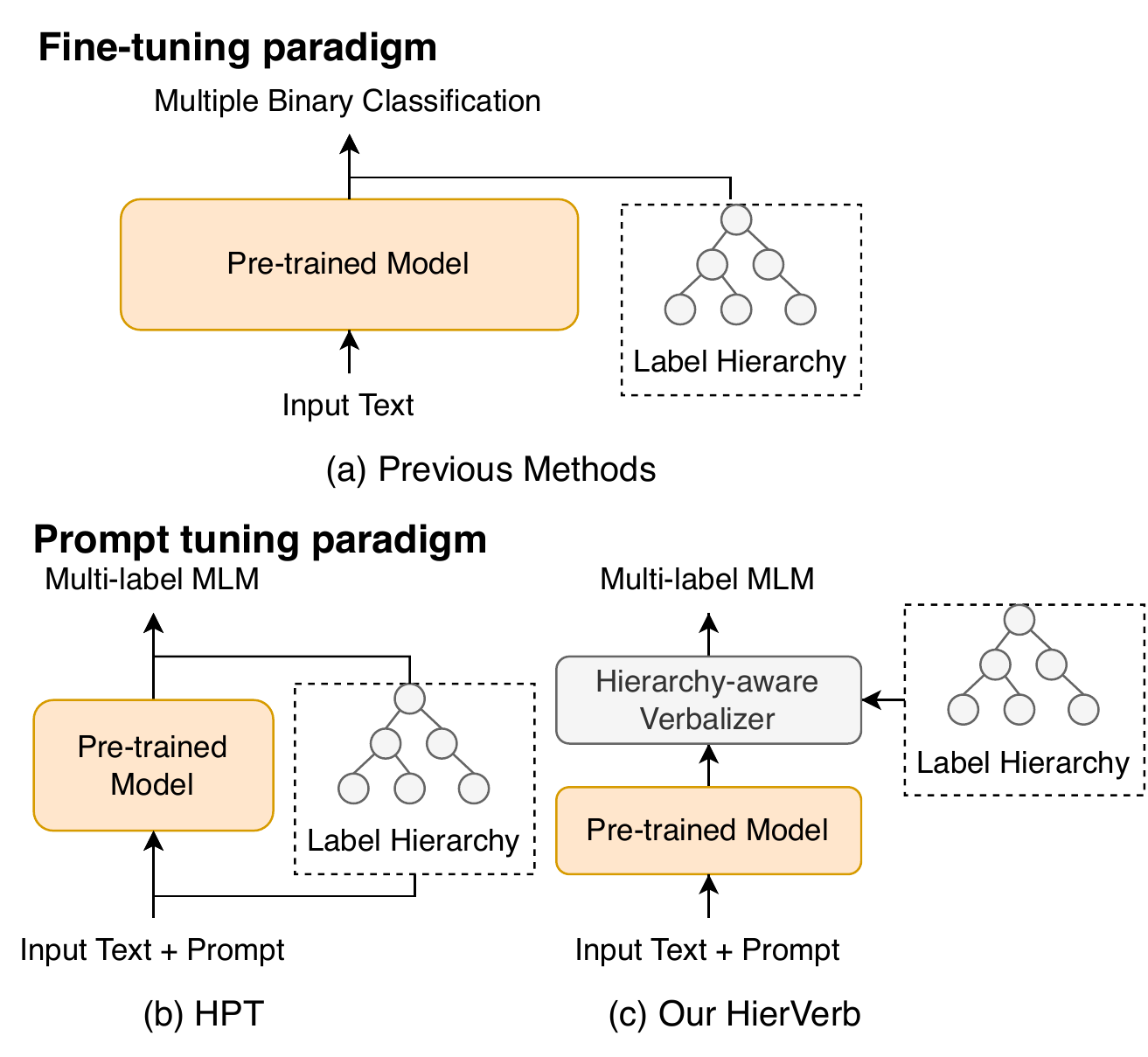}
\caption{Illustration of methods for HTC problems. (a) Previous methods typically regard HTC as a downstream classification fine-tuning task. (b) HPT ~\citep{wang2022hpt} formulates HTC as a multi-label MLM problem following the prompt tuning paradigm. (c) Our HierVerb leverages hierarchy-aware verbalizers, which are more effective for few-shot tuning.}
\label{fig:comparison with hpt}
\end{figure}

Existing state-of-the-art HTC models focus on inserting label hierarchy features through graph encoders and then fuse the features into the input layer \citep{wang2022hpt} or output layer \citep{zhou-etal-2020-hierarchy} of a text encoder such as Bidirectional LSTM or pre-trained language models (PLMs), as shown in Figure \ref{fig:comparison with hpt}(a). And there is a trend of taking advantage of PLMs \citep{chen-etal-2021-hierarchy,wang2022hpt} as the backbone of the text encoder through a fine-tuning paradigm. Despite the success of PLMs \citep{devlin-etal-2019-bert,raffel2020exploring} in extensive NLP-related tasks, recently, a series of studies \citep{petroni-etal-2019-language,davison-etal-2019-commonsense,chen2022knowprompt} suggest that it's helpful to elicit the knowledge contained in PLMs and point out the fine-tuning paradigm is suboptimal in few-shot settings due to distinct training strategies between the pre-training and fine-tuning stages. Inspired by "in-context learning" proposed by GPT-3 \citep{NEURIPS2020_1457c0d6}, lots of prompt-based \citep{petroni-etal-2019-language,gao-etal-2021-making,schick-schutze-2021-exploiting,qin-eisner-2021-learning} methods were proposed to bridge the gap between pre-training and downstream tasks via stimulating pre-trained model knowledge with a few hard or soft prompts. 
In prompt-based tuning, the input is usually wrapped through a natural language template and the tasks are converted as masked language modeling (MLM) for PLM.
For example, in the sentiment classification task, the original input x will be wrapped as "x. It was \texttt{[MASK]}”.
The objective is to utilize MLM to predict the word that fills the \texttt{[MASK]}, and subsequently employ a \textit{verbalizer} to map the predicted word to the final classification (e.g. "positive" -> label "Positive").

Although remarkable performances have been achieved via prompt tuning on the flat text classification where labels have no hierarchy, its effects on HTC problems remain unclear, as discussed in HPT \citep{wang2022hpt}.
As shown in Figure \ref{fig:comparison with hpt}(b), HPT proposed a hierarchy-aware prompt tuning method that incorporates the label hierarchy knowledge into soft prompts through graph representation and achieves the new state-of-the-art results on several HTC popular datasets. However, even though the low-resource setting experiment was considered in HPT, the commonly used K-shot setting was not investigated. 
The limitation lies in the absence of a uniform definition of the K-shot setting in HTC. 
Besides, the way to utilize PLMs in few-shot settings through soft prompts and fuse hierarchy by graph encoder into the PLMs harms tapping the full potential of PLMs. 
Hence, it is crucial to exploit a new method to elicit knowledge from PLMs in a hierarchy-aware manner for few-shot learning. 

Inspired by the prior works on verbalizer design \citep{gao-etal-2021-making, schick-schutze-2021-exploiting} between model outputs and labels, as shown in Figure \ref{fig:hierverb2}(a) and \ref{fig:hierverb2}(b),  which makes promising improvements over prompt-based tuning, it is natural to raise this question: is there any verbalizer design method specific to the HTC problems? 
The most current works can be mainly divided into three kinds of verbalizers: manual verbalizers, search-based verbalizers, and soft verbalizers.
However, the main difference between previous works on verbalizers is the way of embedding the semantic space and they are all based on a strong assumption that there is no hierarchical dependency between downstream task labels,  which raises a gap between rich flat prior knowledge in PLM and downstream task hierarchies.
Thus these verbalizers are not suitable for hierarchical classification tasks, lacking awareness of hierarchy in their architectural design.
To address these issues, we introduce a hierarchical-aware verbalizer (HierVerb) combined with the prompt tuning method to fully exploit the hierarchical knowledge within labels. The major contributions of this paper can be summarized as follows:

\begin{itemize}
     \item To our best knowledge, we are the first to define the path-based few-shot setting on hierarchical text classification tasks and propose a path-based evaluation metric to further explore the consistency problem in HTC tasks.
     \item We propose HierVerb for few-shot HTC, which integrates the hierarchical information into the verbalizers through the flat hierarchical contrastive learning and hierarchy-aware constraint chain to better leverage the pre-trained language model for few-shot learning.
     \item Experimental results demonstrate that HierVerb significantly outperforms the current state-of-the-art HTC methods on three popular benchmarks (WOS, DBPedia, and RCV1-V2) under extreme few-shot settings (i.e., K <=8), validating the effectiveness of its design.
\end{itemize}


\section{Related Work}

\subsection{Hierarchical Text Classification}
Current works for HTC focus on finding ways to insert the hierarchical label knowledge into the model, which proves to be beneficial for the problem induced by the imbalanced and large-scale label hierarchy faced in HTC problems \citep{mao-etal-2019-hierarchical}.
Several works \citep{zhang2022hcn,wu-etal-2019-learning,mao-etal-2019-hierarchical} applied the label-based attention module or utilized the meta-learning and reinforcement learning methods to leverage the label structure.
However, as pointed out in HiAGM \citep{zhou-etal-2020-hierarchy}, such methods mainly concentrate on optimizing decoding results based on the constraint of hierarchical paths, it proposed to encode the holistic label structure with hierarchy encoders (graph or tree structure) which demonstrate to improve performance to a greater extent. Following the line of this research, \citet{chen-etal-2021-hierarchy} exploited the relationship between text and label semantics using matching learning, and \citet{wang-etal-2021-concept} explicitly enriched the label embedding with concepts shared among classes. Yet since the label hierarchy representation remains unchanged regardless of the input, later works like HGCLR \citep{wang-etal-2022-incorporating} and HPT \citep{wang2022hpt} chose to migrate label hierarchy into text encoding instead of separately modeling text and labels. In addition to this, HPT achieves state-of-the-art by exploiting pre-trained language models through prompt tuning methods. Although the methods above are designed for HTC problems and prompt-based techniques are applied, the frequently faced few-shot issues in HTC are less investigated, not to mention a suitable solution working well on limited training samples in a hierarchy-aware manner. 

\subsection{Prompt Tuning}
Recent years have observed the widespread and powerful use of pre-trained language models (PLMs) in various downstream NLP tasks \citep{devlin-etal-2019-bert, qiu2020pre, han2021pre}.
Prompt engineering goes a step further by designing a prompt template to take the power of PLMs to unprecedented heights, especially in few-shot settings \citep{liu2021pre}.
Later works focus on automatically discovering better hard prompts described in a discrete space to use in the querying process \citep{jiang-etal-2020-know, gao-etal-2021-making}.
Besides, there come with many methods that learn continuous soft prompts directly in the feature space of PLMs \citep{li-liang-2021-prefix, lester-etal-2021-power, qin-eisner-2021-learning}.
Such continuous prompts reduce the hassle of constructing template words and transform them into parameterized embeddings. 

\subsection{Verbalizer Design}

Verbalizers aim to reduce the gap between model outputs and label words, which has always been a critical issue in prompt-based tuning. Most of the current works leverage human written verbalizers \citep{schick-schutze-2021-exploiting} that prove to be effective to build bridges between them.
However, these approaches are highly biased towards lexical semantics of manual verbalizers and require both domain expertise of downstream tasks and understanding of the PLMs' abilities \citep{schick-etal-2020-automatically}.
\citet{schick-etal-2020-automatically} and other studies \citep{gao-etal-2021-making, shin-etal-2020-autoprompt} have designed search-based verbalizers for better verbalizer choices during the training optimization process, intending to reduce the bias caused by personal vocabulary and the cost of intensive human labor.
Another line of researches \citep{hambardzumyan-etal-2021-warp, cui-etal-2022-prototypical} claims it is hard to find satisfactory label words by searching large vocabulary with few examples and proposes to insert learnable embedding vectors as soft labels/verbalizers optimized during the training process. Nevertheless, the verbalizer design methods for hierarchical labels are less explored in previous works.
\begin{figure}[!t]
\centering
\includegraphics[width=0.47\textwidth]{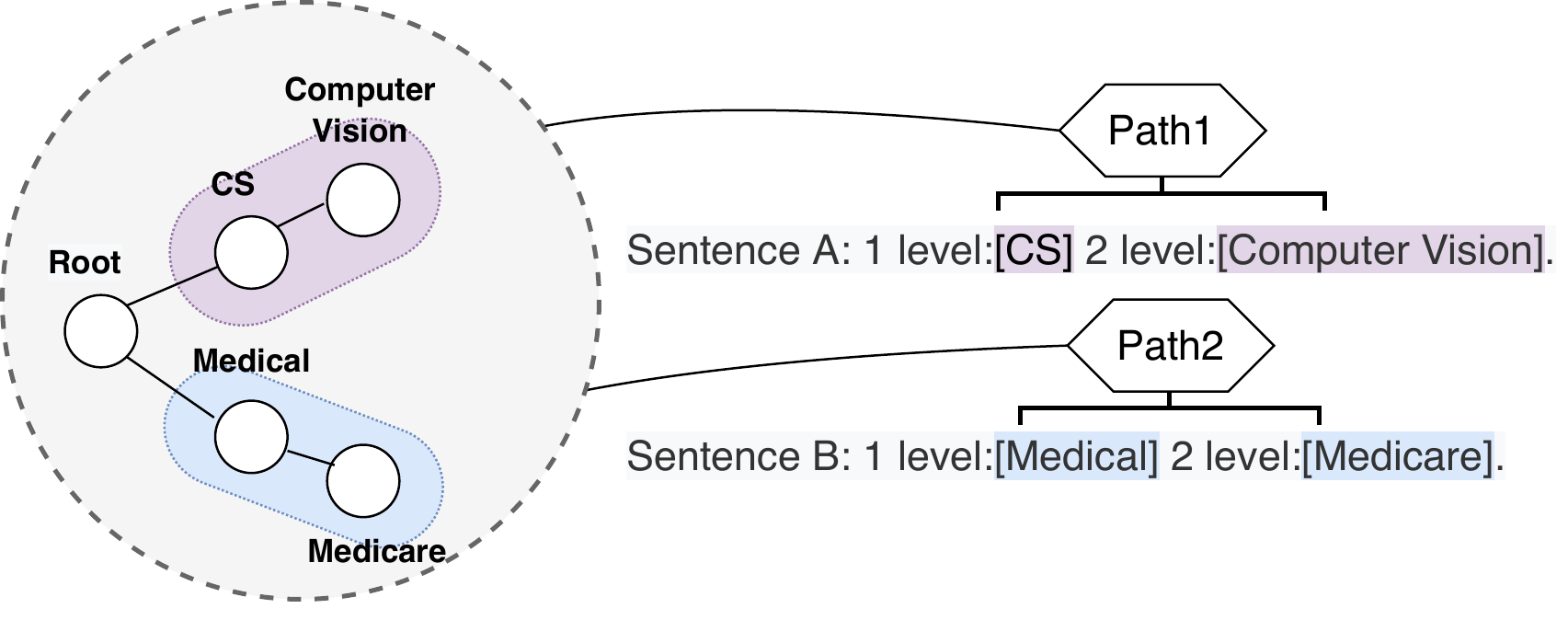}
\caption
{
A few-shot HTC example.
Path 1 contains labels (nodes) for \textit{CS} and \textit{Computer Vision} while Path 2 contains \textit{Medical} and \textit{Medicare}.
Suppose we sample based on the hierarchical label paths in this figure, the sub-dataset consisting of sentences A and B is the support set of our 1-shot HTC.
}
\label{fig:fewshot_settings}
\end{figure}
\section{Preliminaries}

\subsection{Traditional HTC}
In traditional HTC task, the structure of candidate labels $\textit{y}_i\in \textit{Y}$ are predefined as a Directed Acyclic Graph (DAG) $\mathcal{H}$ = (\textit{Y}, \textit{E}), where \textit{Y} is the label set and \textit{E} denotes the hierarchical connections within the labels. 
Specifically,
$\mathcal{H}$ is a tree-like structure where every node except the root has one and only one parent.
Hence the predicted hierarchical labels for one input sample correspond to single- or multi-path taxonomic labels in $\mathcal{H}$.
It is worth noting that the HTC task is often viewed as a multi-label problem. 
Therefore the standard HTC task can be defined as follows: given an input text \textbf{x}=${\{\textit{x}_t\}}_{t=1}^T$ and a label set \textit{Y}, HTC aims to find a subset \textit{y} from \textit{Y}, in other words, to find one label path or multiple paths in $\mathcal{H}$, for \textbf{x}.

\subsection{Few-shot HTC}
The few-shot problem has been extensively studied on tasks such as text classification, image segmentation, and named entity recognition (NER), while few works focus on the few-shot HTC task, which we call Few-HTC. 
It is easy to perform sampling strategies in flat single-label text classification to select K examples for each class added to the support set of K-shot learning. However, this sampling method is difficult to directly apply to HTC because an input sample may contain multiple labels. Hence it is harder to strictly meet the requirement of K shots for each corresponding class \citep{ding-etal-2021-nerd}.

Inspired by the few-shot settings in named entity recognition \citep{yang-katiyar-2020-simple, ding-etal-2021-nerd}
where they regard entity types as basic classes and sample few-shot sets based on each class through greedy sampling algorithms,
we can define our few-shot settings based on the label paths in $\mathcal{H}$ since multiple slots in NER are analogous to multiple label paths in HTC. Figure~\ref{fig:fewshot_settings} shows how we perform path-based sampling for building a Few-HTC support set.

Formally, the task of K-shot HTC is defined as follows: given a text \textbf{x}=${\{\textit{x}_t\}}_{t=1}^T$ and a K-shot support set \textit{S} for the target mandatory-leaf \citep{NIPS2012_f899139d} path set $\textit{C}_\mathcal{T}$, the goal is to predict all golden paths on the label hierarchy tree for \textbf{x}. We design a greedy sampling method specifically for HTC problems and the details of obtaining $\textit{C}_\mathcal{T}$ and the support set \textit{S} from the original HTC datasets are shown in Algorithm \ref{alg:few-shot} to make sure each label path has exactly K-shot examples.
To the best of our knowledge, we are the first to apply path-based few-shot settings on the HTC tasks.

\begin{algorithm}[t]
\small
\caption{Greedy sampling for Few-shot HTC} 
\label{alg:few-shot}
\hspace*{0.02in} {\bf Input:} 
shot K, original HTC dataset X\{(x,y)\}\ with label hierarchy $\mathcal{H}$ \\
\hspace*{0.02in} {\bf Output:} 
K-shot support set S after sampling
\begin{algorithmic}[1]
\State $\rm \textit{C}_\mathcal{T}\gets //Initialize\ the\ original\ set\ of\ mandatory-leaf\ paths$
\While{$ori\_length \neq cur\_length$}
\State $\rm ori\_length \gets //Obtain\ the\ length \ of\ \textit{C}_\mathcal{T}$
\State $\rm Count\ the\ frequency\ of\ each\ \textit{C}_\textit{i}\ in\ X$
\State $\rm Remove\ paths\ \{\textit{C}_\textit{i}\}\ with\ frequency\ less\ than\ K$
\State $\rm Remove\ samples\ containing\ \{\textit{C}_\textit{i}\} \ in\ X$
\State $\rm cur\_length \gets //Obtain\ the\ length \ of\ \textit{C}_\mathcal{T}$
\EndWhile
\State $\rm \{\textit{C}_\textit{i}:\textit{A}_\textit{i}\} \gets // Count\ the\ frequency\ of\ each\ \textit{C}_\textit{i}$
\Statex $\rm appeared\ individually\ in\ the\ filtered\ dataset\ X$
\State $\rm Sort\ the\ path\ set\ \textit{C}_\mathcal{T}\ based\ on\ A$
\State $\rm S \gets \phi //Initialize\ an\ empty\ support\ set$
\State $\rm \{Count_\textit{i}\} \gets \ //Initialize\ the\ counts\ of\ all\ paths\ in$
\Statex $\rm \textit{C}_\mathcal{T}\ to \ zero$
\For{$i=1\ to\ |C_\mathcal{T}|$}
\While{$\rm Count_\textit{i}<K$}
\State $\rm Sample(x,y) \in X_{s.t.}\textit{C}_\textit{i}\in y, w/o\ replacement$
\State $\rm \textit{S}\gets \textit{S} \cup \{ (x,y)\}$
\State $\rm Update\ \{ Count_\textit{j}\} \forall \ \textit{C}_\textit{j} \in y $
\EndWhile
\EndFor
\State $\textbf{return}\ \textit{S}$
\end{algorithmic}
\end{algorithm}

\begin{figure*}[htbp]
\centering
\includegraphics[width=1\textwidth]{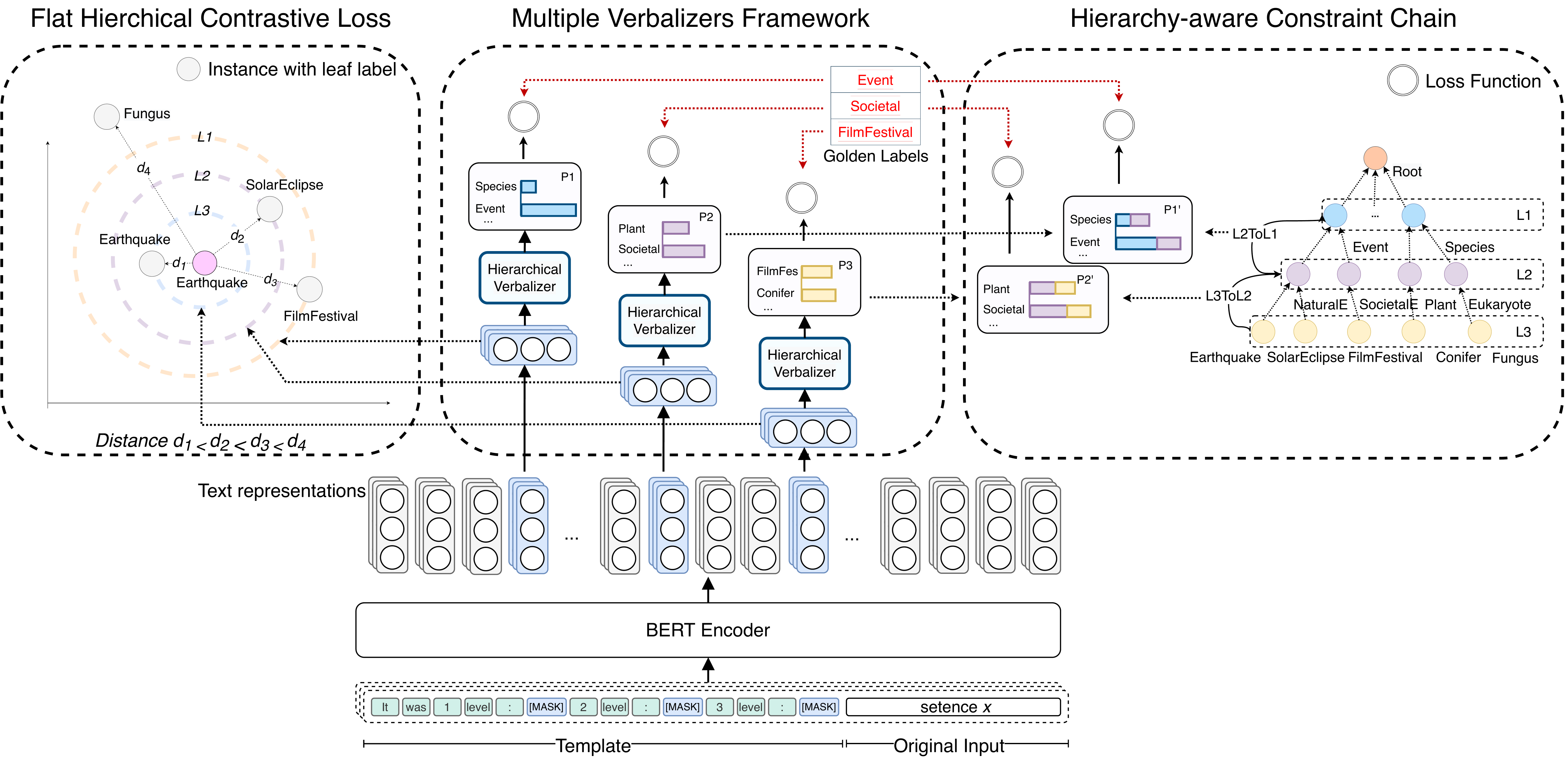}
\caption{
The architecture of HierVerb. There are two ways to insert hierarchical constraints into the model: (a) Hierarchy-aware Constraint Chain. (b) Flat Hierarchical Contrastive Loss. 
P1 to P3 denote the predicted probabilities of labels in each layer. P1$^{\prime}$ and P2$^{\prime}$ represent the propagated probabilities in each layer. L3 to L1 contain labels from a lower to a higher layer. 
$d_1$ to $d_4$ stand for the distances between different input instances. 
Worth noting that the instance with the leaf label "Solar Eclipse" positioned in the L2 circle, for example, shares the same gold labels in layer 2 with the instance marked pink in the center of the circle, and $d_2$ reflects the similarity distance between their output token embeddings of mask tokens corresponding to layer 2. The same applies to other instances.
}
\label{fig:hierverb}
\end{figure*}

\section{Hierarchical Verbalizer}
In this section, we will introduce the proposed hierarchy-aware verbalizer in detail.
We incorporate hierarchical information through our multi-verbalizer framework with prompt templates to elicit rich prior knowledge within the PLMs.
Figure~\ref{fig:hierverb} shows the overall architecture of our proposed HierVerb. We first obtain the hidden states of the multiple mask tokens to represent the sentence and then project it to the verbalizer's space of different label layers.

\subsection{Multi-verbalizer Framework}

Since the label hierarchy is a tree structure in our problem, we think of HTC as a single-label or multi-label classification task performed at multiple levels, following \citet{wang2022hpt}.
In this way, we can easily construct templates based on the depth of the hierarchy tree.
Given a piece of training text \textit{x} and the label hierarchy $\mathcal{H}$ with a depth of $\textit{D}$, the template p is written simply as "\texttt{[CLS] It was 1 level:[MASK] 2 level:[MASK]…\textit{D} level:[MASK]. {\rm \textit{x}} [SEP]}".
We use multiple \texttt{[MASK]} tokens for corresponding multi-level label predictions.
Note that the number of $\texttt{[MASK]}$ tokens is equal to the number of layers of $\mathcal{H}$.

For better learning of hierarchical verbalizer and text representation in few-shot settings, we use \textbf{BERT} \citep{devlin-etal-2019-bert} as our text encoder.
For an input text \textit{x} wrapped with the template \textit{T}:
\begin{equation}
T_{prompt}(x) = \text{\{[CLS] It was $t_i$ ... $t_D$. x [SEP]\}}
\end{equation}
where $t_i$ means "\texttt{i level:[MASK]}".
Note that our template \textit{T} is a dynamically wrapped sentence containing as many \textit{t} as the number of hierarchy layers.
We feed the input \textit{x} wrapped with the template \textit{T} to the encoder of the \text{BERT} to obtain the hidden states $\textit{h}_{1:n}$:
\begin{equation}
    h_{1:n} = \text{BERT}(T_{prompt}(x)_{1:n})
\end{equation}
where $h_{1:n} \in \mathbb{R}^{n \times r}$ and \textit{r} is the hidden state dimension of BERT and \textit{n} is the length of $\textit{T}_{prompt}(x)$.
For convenience, we pick out a subset $\{h^d\}(d\in [1,..., D])$ which is the set of hidden state vectors corresponding to all \texttt{[MASK]} tokens.

On top of this, we use multi-verbalizer for depth-oriented learning and construct each verbalizer based on the full set of labels for the corresponding layer. 
Thus we have a list of verbalizers V = \{$V_d$\}$(d\in [1,...,D])$.
Each verbalizer is created as a virtual continuous vector $W_d \in \mathbb{R}^{r \times l_d}$ where $l_d$ is the number of labels of \textit{d}-th layer and we initialize the embedding $W_d$ of each $V_d$ by averaging the embeddings of its corresponding label tokens and label tokens of all its descendants in $\mathcal{H}$.

In our framework, the \textit{d}-th mask is connected to the \textit{d}-th verbalizer to play the role of predicting the \textit{d}-th layer label.

We denote the distribution of the wrapped sentences in the corpus as $\mathcal{O}$.
The probability distribution of all labels $y_d$ on the layer \textit{d} is:
\begin{equation}
    P_\mathcal{O}(y_d|T_{prompt}(x), \mathcal{D}=d)=q(h^d W_d  + b_d)
\end{equation}
where $W_d \in \mathbb{R}^{r \times l_d}$ and $b_d \in \mathbb{R}^{l_d}$ are weights and bias and \textit{q} is a function used to convert logits into probabilities. 
Hence the predicted probability of text \textit{i} on label \textit{j} of \textit{d}-th layer is:
\begin{equation}
    p_{ij}^d = P_\mathcal{O}(y_d=j|T_{prompt}(x), \mathcal{D}=d)
\end{equation}
Following previous work \cite{zhou-etal-2020-hierarchy, wang-etal-2022-incorporating}, we use a binary cross-entropy loss function for multi-label classification. 
However, the definition of multi-label in our framework is slightly different from these works.
The multi-label problem whose ground truth is a single path on the hierarchical dependency tree $\mathcal{H}$ can be redefined as a single-label prediction problem at each layer with the help of the multi-verbalizer.
For such a single-path prediction, the loss function is defined as:
\begin{equation}
L_{idj}^C= -y_{ij}^dlog(p_{ij}^d)
\end{equation}
Instead, for multi-path problems:
\begin{equation}
L_{idj}^C= -y_{ij}^dlog(p_{ij}^d)-(1-y_{ij}^d)log(1-p_{ij}^d)
\end{equation}

To sum up, for each input text \textit{i}, we can calculate the loss of the multi-verbalizer framework as:
\begin{equation}
\mathcal{L}_C = \sum\limits_d^\textit{D}\sum\limits_j^{l_d} L_{idj}^C = \sum\limits_d^\textit{D}\sum\limits_j^{l_d} L^C(p_{ij}^d, y_{ij}^d)
\end{equation}

\subsection{Hierarchy-aware Constraint Chain}
In order to reduce the gap between the training objective of the pre-trained model and the hierarchical objective, we first use the hierarchical constraint chain to solve this problem. 

According to the label dependency tree $\mathcal{H}$, we maintain a parent-to-child mapping $\overrightarrow{M}$ between layers:
\begin{equation}
    \overrightarrow{M}_{d}(y_j^{d}) = \{y_1^{d+1}, y_2^{d+1},...,y_n^{d+1}\}
\end{equation}
where \textit{$y_d$} is a label \textit{j} belonging to the \textit{d}-th layer and $\{y_n^{d+1}\}$ are its corresponding children nodes at the (\textit{d}+1)-th layer.

Thus the propagated probability of text \textit{i} on label \textit{j} of \textit{d}-th layer can be obtained through:
\begin{equation}
    \tilde{p}_{ij}^d = (1-\beta) p_{ij}^{d} + \beta \sum p_{i\tilde{j}}^{d+1}, \tilde{j}\in \overrightarrow{M}_d(j)
\end{equation}
which is implemented to quantify constraints from descendant nodes where $\beta$ controls the degree of descendant node constraints. 
Since we are propagating from the bottom up, our computational constraints gradually propagate upward from the leaf nodes of the hierarchy tree.
The loss of the constraint chain can be defined as:
\begin{equation}
    \mathcal{L}_{HCC} = \sum\limits_d^\textit{D}\sum\limits_j^{l_{d-1}} L^C(\tilde{p}_{ij}^d, y_{ij}^d)
\end{equation}

\subsection{Flat Hierarchical Contrastive Loss}
Secondly, we design the flat hierarchical contrastive loss objective to learn the hierarchy-aware matching relationship between instances, instead of the relationship between instances and labels as proposed in \citet{chen-etal-2021-hierarchy}.
It is non-trivial to match different instances due to the sophisticated semantics of each instance in the hierarchical setting.
Given input sentence representation and the label hierarchy, there are two main goals we want to achieve through optimization: 
(1) For sentence pairs, the representations of intra-class correspondences at each level should obtain higher similarity scores than inter-class pairs. 
(2) The similarity between lower-level representations of intra-class pairs deserves more weight than that of relatively high-level ones.
To achieve our goal, we flatten the hierarchy into a multi-level lattice structure and define our objective function based on the SimCSE estimator \citep{gao-etal-2021-simcse}, which is widely used in contrastive learning.

\begin{table}[b]
\begin{center}
\resizebox{\linewidth}{!}{
\begin{tabular}{l|llll}
\hline
 &DBPedia &WOS &RCV1-V2 \\ 
\hline
Level 1 Categories & 9 & 7 &4 \\
Level 2 Categories & 70 & 134 &55 \\
Level 3 Categories & 219 & NA &43 \\
Level 4 Categories & NA & NA &1\\
Number of documents & 381025 & 46985 &804410\\
Mean document length &106.9 &200.7 &221.29\\ 
\hline
\end{tabular}
}
\end{center}
\caption{Comparison of popular HTC datasets.}
\label{tab:dataset}
\end{table}

Denote $\textbf{B}=\{(X_{n}, \{Y^d\}_n) \}$ as one batch where $\{{Y^d}\}_n$ is the original labels in $\textit{d}$-th layer,
$\textit{n}\in \textit{N}$, $d \in \textit{D}$, where $\textit{N}$ denotes the batch size and $\textit{D}$ denotes the maximum depth of the label hierarchy $\mathcal{H}$.
Following SimCSE, we can have 2N sets of hidden vectors for all corresponding \texttt{[MASK]} tokens $\textbf{Z}=\{z \in \{h^d\} \cup \{\tilde{h}^d \} \}$ where $\tilde{h}^d$ is simply obtained by feeding original text into the encoder for the second time. Any sentence pairs in one batch can be defined as
$\mathcal{P}=[(X_{a}, \{{Y^d}\}_a), (X_{b}, \{{Y^d}\}_b)]$, and we keep a lattice label matrix:
\begin{equation}
M_{d}(a, b)= \left \{
\begin{array}{ll}
    1, & \{{Y^d}\}_a \cap \{{Y^d}\}_b \neq \phi\\
    0, & \{{Y^d}\}_a \cap \{{Y^d}\}_b = \phi
\end{array}
\right.
\end{equation}

Thus the final flat hierarchical contrastive loss function is defined as:
\begin{equation}
\resizebox{.9\hsize}{!}{$
    L_{\textit{\texttt{FHC}}}=\frac{-1}{N^2D^2}
        \sum\limits_d^D
        \sum\limits_u^d
        \sum\limits_n^{2N}
        \log
        \frac{\exp(\sum_{n'}   S(h_n^u, h_{n'}^u) M_u(n, n'))}
             {\exp(\sum_{n'}   S(h_n^u, h_{n'}^u))}
        \times \frac{1}{2^{(D-d)\times\alpha}}
$
}
\end{equation}
where \textit{S} is cosine similarity function, $\textit{h}_n^d$ is the hidden states of \textit{d}-th \texttt{[MASK]} for sentence \textit{n}, and $\alpha$ controls the relative penalty importance of different layers. 

Considering that once $M_d(n, n')$ equals to one, all $M_u(n, n')$ can be assured to be equal to one by reason of the tree structure. 
Thereafter it assigns more weight to the contrastive loss of the lower layer whose \textit{d} value is larger, and $\alpha$ intensifies the differentiation between all layers. This results in the inequality $Distance\ d_1<d_2<d_3<d_4$ in Figure \ref{fig:hierverb}.


\subsection{Classification Objective Function}
Overall, our final training objective is the combination of multi-verbalizer framework loss, constraint chain loss, and flat hierarchical contrastive loss.
\begin{equation}
    \mathcal{L}=\mathcal{L}_C+\lambda_1 \mathcal{L}_{HCC}+\lambda_2 \mathcal{L}_{FHC}
\end{equation}
where $\lambda 1$ and $\lambda 2$ are the hyperparameters controlling the weights of corresponding loss and HCC and FHC stand for Hierarchy-aware Constraint Chain and Flat Hierarchical Contrastive Loss respectively.

\begin{table*}[!ht]
	\centering
	\tiny
	\setlength{\tabcolsep}{1.2mm}
	\begin{tabular}{lccccccc}
		\toprule[1pt]
		\multicolumn{1}{c}{\multirow{2}{*}
		    {\begin{tabular}[c]{  @{}c@{}} \\ K\end{tabular}}}
		 & \multirow{2}{*}
		    {\begin{tabular}[c]{@{}c@{}} \\ Method\end{tabular}} 
		 
         & \multicolumn{2}{c}{\textbf{WOS(Depth 2)}} 
         & \multicolumn{2}{c}{\textbf{DBpedia(Depth 3)}} 
         & \multicolumn{2}{c}{\textbf{RCV1-V2(Depth 4)}}\\ \cmidrule {3-8} 
	     \multicolumn{1}{l}{} &  & Micro-F1 & Macro-F1 & Micro-F1  & Macro-F1 & Micro-F1 & Macro-F1  \\ \midrule
	     
		\multicolumn{1}{c}{\multirow{5}{*}{1}} 
		 & \multicolumn{1}{|l}{BERT (Vanilla FT)} 
            &2.99 $\pm$ 20.85 \textcolor[HTML]{387230}{(5.12)} 
            &0.16 $\pm$ 0.10 \textcolor[HTML]{387230}{(0.24)} 
            &14.43 $\pm$ 13.34 \textcolor[HTML]{387230}{(24.27)} 
            &0.29 $\pm$ 0.01 \textcolor[HTML]{387230}{(0.32)} 
            &7.32 $\pm$ 10.33 \textcolor[HTML]{387230}{(9.32)} 
            &\underline{3.73} $\pm$ 0.10 \textcolor[HTML]{387230}{(3.73)} 
            \\
		 & \multicolumn{1}{|l}{HiMatch \citep{chen-etal-2021-hierarchy}} 
            &43.44 $\pm$ 8.90 \textcolor[HTML]{387230}{(48.26)}      
            &7.71 $\pm$ 4.90 \textcolor[HTML]{387230}{(9.32)}      
            &-   &-  &-  &-  \\
		 & \multicolumn{1}{|l}{HGCLR\citep {wang-etal-2022-incorporating}} 
            &9.77 $\pm$ 11.77 \textcolor[HTML]{387230}{(16.32)}     
            &0.59 $\pm$ 0.10 \textcolor[HTML]{387230}{(0.63)}     
            &15.73 $\pm$ 31.07 \textcolor[HTML]{387230}{(25.13)}      
            &0.28 $\pm$ 0.10 \textcolor[HTML]{387230}{(0.31)}     
            &26.46 $\pm$ 1.27 \textcolor[HTML]{387230}{(26.80)}     
            &1.34 $\pm$ 0.93 \textcolor[HTML]{387230}{(1.71)}   
            \\
		 & \multicolumn{1}{|l}{HPT \citep{wang2022hpt}} 
            &\underline{50.05} $\pm$ 6.80 \textcolor[HTML]{387230}{(50.96)}   
            &\underline{25.69} $\pm$ 3.31 \textcolor[HTML]{387230}{(27.76)} 
            &\underline{72.52} $\pm$ 10.20 \textcolor[HTML]{387230}{(73.47)}  
            &\underline{31.01} $\pm$ 2.61 \textcolor[HTML]{387230}{(32.50)} 
            &\underline{27.70} $\pm$ 5.32 \textcolor[HTML]{387230}{(28.51)}   
            &3.35 $\pm$ 2.22 \textcolor[HTML]{387230}{(3.90)}   
            \\ \cmidrule {2-8}
         & \multicolumn{1}{|l}{HierVerb} 
         &\textbf{58.95} $\pm$ \textbf{6.38} \textcolor[HTML]{387230}{(\textbf{61.76})}  &\textbf{44.96} $\pm$ \textbf{4.86}  \textcolor[HTML]{387230}{(\textbf{48.19})}   &\textbf{91.81} $\pm$ \textbf{0.07} \textcolor[HTML]{387230}{(\textbf{91.95})}  &\textbf{85.32} $\pm$ \textbf{0.04} \textcolor[HTML]{387230}{(\textbf{85.44})}  &\textbf{40.95} $\pm$ \textbf{3.12} \textcolor[HTML]{387230}{(\textbf{41.22})}  &\textbf{4.87} $\pm$ \textbf{1.71} \textcolor[HTML]{387230}{(\textbf{5.71})} \\ \midrule
        
        \multicolumn{1}{c}{\multirow{5}{*}{2}} 
         & \multicolumn{1}{|l}{BERT (Vanilla FT)} 
         &46.31 $\pm$ 0.65  \textcolor[HTML]{387230}{(46.85)}    
         &5.11 $\pm$ 1.31  \textcolor[HTML]{387230}{(5.51)}  
         &87.02 $\pm$ 3.89  \textcolor[HTML]{387230}{(88.20)}  
         &69.05 $\pm$ 26.81  \textcolor[HTML]{387230}{(73.28)}  
         &8.07 $\pm$ 2.18  \textcolor[HTML]{387230}{(9.13)}  
         &2.76 $\pm$ 6.01  \textcolor[HTML]{387230}{(4.11)}  
         \\
		 & \multicolumn{1}{|l}{HiMatch \citep{chen-etal-2021-hierarchy}} 
         &46.41 $\pm$ 1.31  \textcolor[HTML]{387230}{(47.23)}    
         &18.97 $\pm$ 0.65  \textcolor[HTML]{387230}{(21.06)}  
         &-   &-  &-  &-  \\
	& \multicolumn{1}{|l}{HGCLR \citep{wang-etal-2022-incorporating}} 
         &45.11 $\pm$ 5.02  \textcolor[HTML]{387230}{(47.56)}   
         &5.80 $\pm$ 11.63  \textcolor[HTML]{387230}{(9.63)}    
         &87.79 $\pm$ 0.40  \textcolor[HTML]{387230}{(88.42)}   
         &71.46 $\pm$ 0.17  \textcolor[HTML]{387230}{(71.78)}    
         &34.33 $\pm$ 4.81  \textcolor[HTML]{387230}{(37.28)}     
         &2.51 $\pm$ 6.12  \textcolor[HTML]{387230}{(6.12)}  
         \\
		 & \multicolumn{1}{|l}{HPT \citep{wang2022hpt}} 
         &\underline{57.45} $\pm$ 1.89 \textcolor[HTML]{387230}{(58.99)}    
         &\underline{35.97} $\pm$ 11.89  \textcolor[HTML]{387230}{(39.94)}    
         &\underline{90.32} $\pm$ 0.64  \textcolor[HTML]{387230}{(91.11)}    
         &\underline{81.12} $\pm$ 1.33  \textcolor[HTML]{387230}{(82.42)}   
         &\underline{38.93} $\pm$ 3.55  \textcolor[HTML]{387230}{(40.47)}    
         &\underline{8.31} $\pm$ 5.26  \textcolor[HTML]{387230}{(10.52)}   
         \\ \cmidrule {2-8}
         & \multicolumn{1}{|l}{HierVerb} 
         &\textbf{66.08} $\pm$ \textbf{4.19} \textcolor[HTML]{387230}{(\textbf{68.01})}  &\textbf{54.04} $\pm$ \textbf{3.24} \textcolor[HTML]{387230}{(\textbf{56.69})}   
         &\textbf{93.71} $\pm$ \textbf{0.01} \textcolor[HTML]{387230}{(\textbf{93.87})}  
         &\textbf{88.96} $\pm$ \textbf{0.02} \textcolor[HTML]{387230}{(\textbf{89.02})} 
         &\textbf{48.00} $\pm$ \textbf{2.27} \textcolor[HTML]{387230}{(\textbf{49.21})}  
         &\textbf{11.74} $\pm$ \textbf{1.58} \textcolor[HTML]{387230}{(\textbf{12.69})}  \\ \midrule
        
        \multicolumn{1}{c}{\multirow{5}{*}{4}} 
         & \multicolumn{1}{|l}{BERT (Vanilla FT)} 
         &56.00 $\pm$ 4.25  \textcolor[HTML]{387230}{(57.18)}    
         &31.04 $\pm$ 16.65  \textcolor[HTML]{387230}{(33.77)}   
         &92.94 $\pm$ 0.66  \textcolor[HTML]{387230}{(93.38)}   
         &84.63 $\pm$ 0.17  \textcolor[HTML]{387230}{(85.47)}    
         &17.94 $\pm$ 0.01  \textcolor[HTML]{387230}{(18.00)}    
         &1.45 $\pm$ 0.01  \textcolor[HTML]{387230}{(1.57)}    
         \\
		 & \multicolumn{1}{|l}{HiMatch \citep{chen-etal-2021-hierarchy}} 
         &57.43 $\pm$ 0.01  \textcolor[HTML]{387230}{(57.43)}    
         &39.04 $\pm$ 0.01  \textcolor[HTML]{387230}{(39.04)}    
         &-   &-  &-  &-  \\
		 & \multicolumn{1}{|l}{HGCLR \citep{wang-etal-2022-incorporating}} 
         &56.80 $\pm$ 4.24  \textcolor[HTML]{387230}{(57.96)}    
         &32.34 $\pm$ 15.39  \textcolor[HTML]{387230}{(33.76)}    
         &93.14 $\pm$ 0.01  \textcolor[HTML]{387230}{(93.22)}   
         &84.74 $\pm$ 0.11  \textcolor[HTML]{387230}{(85.11)}  
         &45.53 $\pm$ 4.20  \textcolor[HTML]{387230}{(47.71)}    
         &8.56 $\pm$ 1.63  \textcolor[HTML]{387230}{(9.92)}  
         \\
		 & \multicolumn{1}{|l}{HPT \citep{wang2022hpt}} 
         &\underline{65.57} $\pm$ 1.69  \textcolor[HTML]{387230}{(67.06)}    
         &\underline{45.89} $\pm$ 9.78  \textcolor[HTML]{387230}{(49.42)}     
         &\underline{94.34} $\pm$ 0.28  \textcolor[HTML]{387230}{(94.83)}    
         &\underline{90.09} $\pm$ 0.87  \textcolor[HTML]{387230}{(91.12)}   
         &\underline{52.62} $\pm$ 0.20  \textcolor[HTML]{387230}{(52.73)}
         &\underline{20.01} $\pm$ 0.31  \textcolor[HTML]{387230}{(20.21)}    
         \\ \cmidrule {2-8}
         & \multicolumn{1}{|l}{HierVerb}  
         &\textbf{72.58} $\pm$ \textbf{0.83}  \textcolor[HTML]{387230}{(\textbf{73.64})}      
         &\textbf{63.12} $\pm$ \textbf{1.48}  \textcolor[HTML]{387230}{(\textbf{64.47})}      
         &\textbf{94.75} $\pm$ \textbf{0.13}  \textcolor[HTML]{387230}{(\textbf{95.13})}      
         &\textbf{90.77} $\pm$ \textbf{0.33}  \textcolor[HTML]{387230}{(\textbf{91.43})}      
         &\textbf{56.86} $\pm$ \textbf{0.44}  \textcolor[HTML]{387230}{(\textbf{57.11})}      
         &\textbf{22.07} $\pm$ \textbf{0.32}  \textcolor[HTML]{387230}{(\textbf{22.42})}     
         \\ \midrule
        
        \multicolumn{1}{c}{\multirow{5}{*}{8}} 
         & \multicolumn{1}{|l}{BERT (Vanilla FT)} 
         &66.24 $\pm$ 1.96  \textcolor[HTML]{387230}{(67.53)}    
         &50.21 $\pm$ 5.05  \textcolor[HTML]{387230}{(52.60)}    
         &94.39 $\pm$ 0.06  \textcolor[HTML]{387230}{(94.57)}   
         &87.63 $\pm$ 0.28  \textcolor[HTML]{387230}{(87.78)}   
         &57.27 $\pm$ 0.04  \textcolor[HTML]{387230}{(57.51)}   
         &23.93 $\pm$ 0.45  \textcolor[HTML]{387230}{(24.46)}    
         \\
		 & \multicolumn{1}{|l}{HiMatch \citep{chen-etal-2021-hierarchy}} 
         &69.92 $\pm$ 0.01  \textcolor[HTML]{387230}{(70.23)}    
         &57.47 $\pm$ 0.01  \textcolor[HTML]{387230}{(57.78)}  
         &-   &-  &-  &-  \\
		 & \multicolumn{1}{|l}{HGCLR \citep{wang-etal-2022-incorporating}} 
         &68.34 $\pm$ 0.96  \textcolor[HTML]{387230}{(69.22)}    
         &54.41 $\pm$ 2.97  \textcolor[HTML]{387230}{(55.99)}    
         &94.70 $\pm$ 0.05  \textcolor[HTML]{387230}{(94.94)}   
         &88.04 $\pm$ 0.25  \textcolor[HTML]{387230}{(88.61)}    
         &58.90 $\pm$ 1.61  \textcolor[HTML]{387230}{(60.30)}    
         &27.03 $\pm$ 0.20  \textcolor[HTML]{387230}{(27.41)}  
         \\
		 & \multicolumn{1}{|l}{HPT \citep{wang2022hpt}} 
         &\underline{76.22} $\pm$ 0.99  \textcolor[HTML]{387230}{(77.23)}    
         &\underline{67.20} $\pm$ 1.89  \textcolor[HTML]{387230}{(68.63)}   
         &\underline{95.49} $\pm$ 0.01  \textcolor[HTML]{387230}{(95.57)}    
         &\underline{92.35} $\pm$ 0.03  \textcolor[HTML]{387230}{(92.52)}   
         &\underline{59.92} $\pm$ 4.25 \textcolor[HTML]{387230}{(61.47)}     
         &\underline{29.03} $\pm$ 6.23  \textcolor[HTML]{387230}{(32.19)}   
         \\ \cmidrule {2-8}
         & \multicolumn{1}{|l}{HierVerb} 
        &\textbf{78.12} $\pm$ \textbf{0.55}  \textcolor[HTML]{387230}{(\textbf{78.87})}    
        &\textbf{69.98} $\pm$ \textbf{0.91}  \textcolor[HTML]{387230}{(\textbf{71.04})}    
        &\textbf{95.69} $\pm$ \textbf{0.01}  \textcolor[HTML]{387230}{(\textbf{95.70})}    
        &\textbf{92.44} $\pm$ \textbf{0.01}  \textcolor[HTML]{387230}{(\textbf{92.51})}    
        &\textbf{63.90} $\pm$ \textbf{2.42}  \textcolor[HTML]{387230}{(\textbf{64.96})}    
        &\textbf{31.13} $\pm$ \textbf{1.63}  \textcolor[HTML]{387230}{(\textbf{32.52})}      
        \\ \midrule
        
        \multicolumn{1}{c}{\multirow{5}{*}{16}} 
         & \multicolumn{1}{|l}{BERT (Vanilla FT)} 
         &75.52 $\pm$ 0.32  \textcolor[HTML]{387230}{(76.07)}    
         &65.85 $\pm$ 1.28  \textcolor[HTML]{387230}{(66.96)}    
         &95.31 $\pm$ 0.01  \textcolor[HTML]{387230}{(95.37)}   
         &89.16 $\pm$ 0.07  \textcolor[HTML]{387230}{(89.35)}   
         &63.68 $\pm$ 0.01  \textcolor[HTML]{387230}{(64.10)}    
         &34.00 $\pm$ 0.67  \textcolor[HTML]{387230}{(34.41)}  
         \\
		 & \multicolumn{1}{|l}{HiMatch \citep{chen-etal-2021-hierarchy}} 
         &77.67 $\pm$ 0.01  \textcolor[HTML]{387230}{(78.24)}    
         &68.70 $\pm$ 0.01  \textcolor[HTML]{387230}{(69.58)}  
         &-   &-  &-  &-  \\
		 & \multicolumn{1}{|l}{HGCLR \citep{wang-etal-2022-incorporating}} 
         &76.93 $\pm$ 0.52  \textcolor[HTML]{387230}{(77.46)}  
         &67.92 $\pm$ 1.21  \textcolor[HTML]{387230}{(68.66)}  
         &95.49 $\pm$ 0.04  \textcolor[HTML]{387230}{(95.63)}    
         &89.41 $\pm$ 0.09  \textcolor[HTML]{387230}{(89.71)}    
         &63.91 $\pm$ 1.42  \textcolor[HTML]{387230}{(64.81)}    
         &33.25 $\pm$ 0.10  \textcolor[HTML]{387230}{(33.50)}  
         \\
		 & \multicolumn{1}{|l}{HPT \citep{wang2022hpt}} 
         &\underline{79.85} $\pm$ 0.41  \textcolor[HTML]{387230}{(80.58)}    
         &\underline{72.02} $\pm$ 1.40  \textcolor[HTML]{387230}{(73.31)}    
         &\underline{96.13} $\pm$ 0.01  \textcolor[HTML]{387230}{(96.21)}    
         &\textbf{93.34} $\pm$ \textbf{0.02}  \textcolor[HTML]{387230}{(\textbf{93.45})}   
         &\textbf{65.73} $\pm$ \textbf{0.80}  \textcolor[HTML]{387230}{(\textbf{66.24})}     
         &\textbf{36.34} $\pm$ \textbf{0.20}  \textcolor[HTML]{387230}{(\textbf{36.57})}     
         \\ \cmidrule {2-8}
         & \multicolumn{1}{|l}{HierVerb} 
         &\textbf{80.93} $\pm$ \textbf{0.10}  \textcolor[HTML]{387230}{(\textbf{81.26})}      
         &\textbf{73.80} $\pm$ \textbf{0.12}  \textcolor[HTML]{387230}{(\textbf{74.19})}      
         &\textbf{96.17} $\pm$ \textbf{0.01}  \textcolor[HTML]{387230}{(\textbf{96.21})}      
         &\underline{93.28} $\pm$ 0.06  \textcolor[HTML]{387230}{(93.49)}      
         &\underline{65.50} $\pm$ 1.41  \textcolor[HTML]{387230}{(66.62)}      
         &\underline{35.10} $\pm$ 1.73  \textcolor[HTML]{387230}{(36.24)}     
         \\ 
        
		\bottomrule[1pt]
	\end{tabular}
	\caption{\label{Result} F1 scores on 3 datasets. We report the mean F1 scores ($\%$) over 3 random seeds. \textbf{Bold}: best results. \textbf{Underlined}: second highest. }
\label{tab:main_results}
\end{table*}

\section{Experiments}
\subsection{Experiments Setup}
\setParDis
\paragraph{Experimental settings}
As mentioned in Preliminaries, we focus on few-shot settings that only K samples for each label path are available for training on a new HTC task called Few-HTC in this work.
In order to better study the few-shot generalization ability of the model under different scales of training data, we conduct experiments based on K $\in$ \{1,2,4,8,16\}.
\paragraph{Datasets and Implementation Details}
We evaluate our proposed method on three widely used datasets for hierarchical text classification: Web-of-Science (WOS) \citep{kowsari2017hdltex}, DBpedia \citep{sinha-etal-2018-hierarchical} and RCV1-V2 \citep{lewis2004rcv1}. 
WOS and DBPedia are for single-path HTC while RCV1-V2 includes multi-path taxonomic labels.
The statistic details are illustrated in Table~\ref{tab:dataset}.
For implementation details, please refer to Appendix \ref{sec:imp_details}.

\paragraph{Evaluation Metrics}
Similar to previous work, we measure the experimental results with Macro-F1 and Micro-F1.
To further evaluate the consistency problem between layers, we adopt path-constrained MicroF1 (C-MicroF1) and path-constrained MacroF1 (C-MacroF1) proposed in \citet{yu2022constrained} which we refer to collectively as C-metric.
In C-metric, a correct prediction for a label node is valid only if all its ancestor nodes are correct predictions, otherwise, it is regarded as a misprediction.
However, in the case of path splitting based on the mandatory-leaf nodes, the metric is still not sufficient to provide a comprehensive evaluation of hierarchical path consistency, because it ignores the correctness of a node's children nodes.
Therefore, we propose a new path-constrained evaluation method based on the perspective of path correctness, which is called P-metric (PMacro-F1 and PMicro-F1).
The details of our P-metric are shown in Appendix \ref{sec:path_metric}.

\paragraph{Baselines}
We select a few recent state-of-the-art works as baselines: HiMatch (Using BERT as encoder) \citep{chen-etal-2021-hierarchy}, HGCLR \citep{wang-etal-2022-incorporating} and HPT \citep{wang2022hpt}.
We also perform the vanilla fine-tuning method on the Few-shot HTC task, which we refer to as Vanilla FT in the following.

\begin{table}[!t]
	\centering
	\footnotesize
	\setlength{\tabcolsep}{0.4mm}
	\begin{tabular}{clcccc}
		\toprule[1pt]
		\multicolumn{1}{c}{\multirow{2}{*}{\begin{tabular}[c]{@{}c@{}} \\ 
            K
        \end{tabular}}} & \multicolumn{1}{l}{\multirow{2}{*}{\begin{tabular}[c]{@{}c@{}} \\ Method\end{tabular}}}
		& \multicolumn{4}{c}{\textbf{WOS}} \\ \cmidrule {3-6} 
		& & PMicro-F1 & PMacro-F1   &CMicro-F1 &CMacro-F1 \\ \midrule
  
            \multicolumn{1}{c}{\multirow{4}{*}{1}} 
		& \multicolumn{1}{|l}{Ours} &\textbf{39.77}  &\textbf{37.24} &\textbf{55.18} &\textbf{39.42}  \\
		& \multicolumn{1}{|l}{HPT} &19.97  &17.47 &49.10 &22.92 \\
		& \multicolumn{1}{|l}{HGCLR} &0.0  &0.0 &2.21 &0.09  \\
		& \multicolumn{1}{|l}{Vanilla FT} &0.0  &0.0  &0.96 &0.04 \\
		\midrule
		
		\multicolumn{1}{c}{\multirow{4}{*}{2}} 
		& \multicolumn{1}{|l}{Ours} &\textbf{50.15}  &\textbf{47.98} &\textbf{62.90} &\textbf{49.67} \\
		& \multicolumn{1}{|l}{HPT} &28.27  &26.51 &56.64 &33.50 \\
		& \multicolumn{1}{|l}{HGCLR} &1.39 &1.49 &45.01 &4.88  \\
		& \multicolumn{1}{|l}{Vanilla FT} &1.43 &1.42 &45.75 &4.95  \\
		\midrule
		
		\multicolumn{1}{c}{\multirow{4}{*}{4}} 
		& \multicolumn{1}{|l}{Ours} &\textbf{62.16}  &\textbf{59.70} &\textbf{72.41}  &\textbf{61.19} \\
		& \multicolumn{1}{|l}{HPT} &50.96  &48.76 &69.43 &55.27 \\
		& \multicolumn{1}{|l}{HGCLR} &29.94  &27.70  &57.43  &34.03 \\
		& \multicolumn{1}{|l}{Vanilla FT} &22.97  &20.73  &55.10 &27.50 \\

		\bottomrule[1pt]
	\end{tabular}
	\caption
	{
    	\label{Result} 
    	Consistency experiments on the WOS dataset using two path-constraint metrics. PMicro-F1 and PMacro-F1 are our proposed path-based consistency evaluation P-metric.
            We report the mean F1 scores ($\%$) over 3 random seeds.
            For display, here we call BERT (Vanilla FT) as Vanilla FT.
            \textbf{Bold}: best results.
	}
\label{tab:consistency_wos}
\end{table}

\subsection{Main Results}
\label{sec:main_results}
\setParDef
Main experimental results are shown in Table~\ref{tab:main_results}.
As is shown, HierVerb wins over all comparison models by a dramatic margin under nearly all situations.
Appendix \ref{sec:performance_gap} more intuitively shows the performance gap between different models.

In the case of no more than 4 shots on WOS, 8.9\%, 9.18\%, and 6.91\% micro-F1 absolute improvement and 19.27\%, 18.3\%, and 16.87\% macro-F1 absolute improvement from the best baseline methods are achieved, respectively.
Under 1-shot situations, compared with all baseline models, there is an average of 57.58\% micro, 74.79\% macro-F1 absolute improvement on DBPedia, and 20.46\% micro-F1, 2.06\% macro-F1 absolute improvement on RCV1-V2.
Although the RCV1-V2 dataset provides no label name which has a negative effect on our verbalizer initialization, our method still achieves state-of-the-art on both Micro-F1 and Macro-F1 under almost all few-shot experiments.

There are three main reasons why HierVerb performs better under the few-shot setting: 
(1) Not require additional learning parameters. 
Previous methods like HPT and HGCLR improve the performance by adding extra parameters to the GNN layers, which could lead to overfitting for few-shot settings; 
(2) Multi-Verb is better than the single-flat verb. 
The previous methods are to first stretch the hierarchical label into a flattened one-dimensional space and then do multi-label prediction, more like a normal multi-label classification task with hierarchical dependencies on labels. In contrast, HierVerb advocates preserving the original hierarchical concept in the architecture through a multi-verb framework. 
(3) Our hierarchical loss is optimized from a semantic perspective for better generalization.



\subsection{Consistency Between Multi-layers}
Table~\ref{tab:consistency_wos} further studies the consistency performance.
Since our method is optimized from a semantic perspective, more consideration is given to the potential semantic dependency between different labels rather than directly fitting specific downstream data, our method still maintains excellent consistency performance in the absence of sufficient labeled training corpora.
It is clear that HGCLR and BERT (Vanilla FT) using the direct fitting method only achieve 0 points in PMicro-F1 and PMacro-F1 under the 1 shot setting.
As for HPT, extra graph parameter learning hurts the generalization of PLMs.
The complete experiments and analyses on the other two datasets are shown in Appendix \ref{sec:consistency_experiments}.

\begin{table}[!t]
	\centering
	\footnotesize
	\setlength{\tabcolsep}{0.4mm}
	\begin{tabular}{clcc}
		\toprule[1pt]
		\multicolumn{1}{c}{\multirow{2}{*}{\begin{tabular}[c]{@{}c@{}} \\ K\end{tabular}}}
		 & \multicolumn{1}{l}{\multirow{2}{*}{\begin{tabular}[c]{@{}c@{}} \\ Ablation Models\end{tabular}}}
		& \multicolumn{2}{c}{\textbf{WOS}} \\ \cmidrule {3-4} 
		& & Micro-F1 & Macro-F1  \\ \midrule
		
		\multicolumn{1}{c}{\multirow{5}{*}{1}}
		& \multicolumn{1}{|l}{Ours} &\textbf{58.95}  &\textbf{44.96}  \\
		& \multicolumn{1}{|l}{$r.m.$ FHC loss} &58.13  &44.63  \\
		& \multicolumn{1}{|l}{$r.m.$ HCC loss} &58.26  &44.27   \\
		& \multicolumn{1}{|l}{$+r.m.$ HCC+FHC loss} &58.35  &44.48   \\
		& \multicolumn{1}{|l}{$+r.m.$ multi-verb  (Vanilla SoftVerb)} &56.11  &41.35   \\
		\midrule
		
		\multicolumn{1}{c}{\multirow{5}{*}{2}} 
		& \multicolumn{1}{|l}{Ours} &\textbf{66.08}  &\textbf{54.04}  \\
		& \multicolumn{1}{|l}{$r.m.$ FHC loss} &65.40  &53.89  \\
		& \multicolumn{1}{|l}{$r.m.$ HCC loss} &65.87  &53.94   \\
		& \multicolumn{1}{|l}{$+r.m.$ HCC+FHC loss} &65.23  &53.47   \\
		& \multicolumn{1}{|l}{$+r.m.$ multi-verb (Vanilla SoftVerb)} &62.31  &49.33   \\
		\midrule
		
		\multicolumn{1}{c}{\multirow{5}{*}{4}} 
		& \multicolumn{1}{|l}{Ours} &\textbf{72.58}  &\textbf{63.12}  \\
		& \multicolumn{1}{|l}{$r.m.$ FHC loss} &72.51  &62.70  \\
		& \multicolumn{1}{|l}{$r.m.$ HCC loss} &72.05  &62.52   \\
		& \multicolumn{1}{|l}{$+r.m.$ HCC+FHC loss} &72.22  &62.22   \\
		& \multicolumn{1}{|l}{$+r.m.$ multi-verb (Vanilla SoftVerb)} &69.58  &58.83   \\
		\midrule
		
		\multicolumn{1}{c}{\multirow{5}{*}{8}} 
		& \multicolumn{1}{|l}{Ours} &\textbf{78.12}  &69.98  \\
		& \multicolumn{1}{|l}{$r.m.$ FHC loss} &77.81  &\textbf{70.28}  \\
		& \multicolumn{1}{|l}{$r.m.$ HCC loss} &77.95  &69.80   \\
		& \multicolumn{1}{|l}{$+r.m.$ HCC+FHC loss} &77.88  &69.85   \\
		& \multicolumn{1}{|l}{$+r.m.$ multi-verb (Vanilla SoftVerb)} &75.99  &66.99   \\
		\midrule
		
		\multicolumn{1}{c}{\multirow{5}{*}{16}} 
		& \multicolumn{1}{|l}{Ours} &\textbf{80.93}  &\textbf{73.80}  \\
		& \multicolumn{1}{|l}{$r.m.$ FHC loss} &80.76  &73.54  \\
		& \multicolumn{1}{|l}{$r.m.$ HCC loss} &80.73  &73.69  \\
		& \multicolumn{1}{|l}{$+r.m.$ HCC+FHC loss} &80.92  &73.61 \\
		& \multicolumn{1}{|l}{$+r.m.$ multi-verb (Vanilla SoftVerb)} &79.62  &70.95   \\
		
		\bottomrule[1pt]
	\end{tabular}
	\caption
	{
    	\label{Result} 
    	Ablation experiments on WOS. 
    	$r.m.$ stands for $remove$ and $+r.m.$ stands for $remove$ on the basis of the previous step. 
    	We report the mean F1 scores ($\%$) over 3 random seeds.
            \textbf{Bold}: best results.
	}
\label{tab:ablation}
\end{table}

\subsection{Ablation Study}
The main parts of our work are the multi-verbalizer framework, hierarchy-aware constraint chain, and flat hierarchical contrastive loss.

To illustrate the effect of these parts, we test our model by gradually removing each component of our model at a time by default, as shown in Table~\ref{tab:ablation}.
We implement Vanilla Soft Verbalizer \citep{hambardzumyan-etal-2021-warp} in our own version which we refer to as SoftVerb in the following for convenience.
Similar to HierVerb, the SoftVerb also uses multiple \texttt{[MASK]} tokens, but only uses a single flat verbalizer to map the label.
Compared to SoftVerb which uses a single flat verbalizer, using multi-verbalizer and integrating hierarchical information into the verbalizer of each layer through FHC and HCC leads to better performance.

\subsection{Effects of Model Scales}

In previous experiments like § \ref{sec:main_results}, we show that HierVerb is powerful on bert-base-uncsaed. 
To further study the ability of HierVerb to utilize the prior knowledge of the pre-trained language model, we conduct experiments on bert-large-uncased.
Table \ref{tab:model_scale} demonstrates that HierVerb consistently outperforms all baseline models in all shot settings.
We find that the gap is even significantly larger for HierVerb and all other baseline models compared to using bert-base-uncased.
For example, under 1-shot setting, HierVerb achieves a 27.92\% increase in macro-F1 and an 11.54\% increase in micro-F1, compared with HPT.
But in the case of bert-base-uncased, the improvements of macro-F1 and micro-F1 are 19.27\% and 8.9\% respectively, which further emphasizes that our model is superior to all baseline models in the ability to mine the prior knowledge of the language model, and this effect is more significant when the scale of the language model increases.

\begin{table}[!t]
	\centering
	\footnotesize
	\setlength{\tabcolsep}{0.4mm}
	\begin{tabular}{clcc}
		\toprule[1pt]
		\multicolumn{1}{c}{\multirow{2}{*}{\begin{tabular}[c]{@{}c@{}} \\ K\end{tabular}}}
		 & \multicolumn{1}{l}{\multirow{2}{*}{\begin{tabular}[c]{@{}c@{}} \\ Method\end{tabular}}}
		& \multicolumn{2}{c}{\textbf{WOS}} \\ \cmidrule {3-4} 
		& & Micro-F1 & Macro-F1  \\ \midrule
		
		\multicolumn{1}{c}{\multirow{4}{*}{1}}
		& \multicolumn{1}{|l}{HierVerb} &\textbf{61.29}  &\textbf{47.70}  \\
		& \multicolumn{1}{|l}{HPT} &49.75  &19.78  \\
		& \multicolumn{1}{|l}{HGCLR} &20.10  &0.50   \\
		& \multicolumn{1}{|l}{BERT (Vanilla FT)} &10.78  &0.25   \\
		\midrule
		
		\multicolumn{1}{c}{\multirow{4}{*}{2}} 
		& \multicolumn{1}{|l}{HierVerb} &\textbf{67.92}  &\textbf{56.92}  \\
		& \multicolumn{1}{|l}{HPT} &60.09  &35.44  \\
		& \multicolumn{1}{|l}{HGCLR} &44.92  &3.23   \\
		& \multicolumn{1}{|l}{BERT (Vanilla FT)} &20.50  &0.34   \\
		\midrule
		
		\multicolumn{1}{c}{\multirow{4}{*}{4}} 
		& \multicolumn{1}{|l}{HierVerb} &\textbf{73.88}  &\textbf{64.80}  \\
		& \multicolumn{1}{|l}{HPT} &69.47  &53.22  \\
		& \multicolumn{1}{|l}{HGCLR} &68.12  &52.92  \\
            & \multicolumn{1}{|l}{BERT (Vanilla FT)} &67.44  &51.66   \\
		\midrule
		
		\multicolumn{1}{c}{\multirow{4}{*}{8}} 
		& \multicolumn{1}{|l}{HierVerb} &\textbf{78.56}  &\textbf{71.01}  \\
		& \multicolumn{1}{|l}{HPT} &77.96  &68.26  \\
		& \multicolumn{1}{|l}{HGCLR} &71.48  &56.91   \\
		& \multicolumn{1}{|l}{BERT (Vanilla FT)} &73.98  &62.82   \\
		\midrule
		
		\multicolumn{1}{c}{\multirow{4}{*}{16}} 
		& \multicolumn{1}{|l}{HierVerb} &\textbf{82.09}  &\textbf{75.01}  \\
		& \multicolumn{1}{|l}{HPT} &80.69  &72.51  \\
		& \multicolumn{1}{|l}{HGCLR} &78.01  &67.87   \\
		& \multicolumn{1}{|l}{BERT (Vanilla FT)} &78.52  &69.64   \\
		
		\bottomrule[1pt]
	\end{tabular}
	\caption
	{
    	\label{Result} 
             Using the same hyperparameter settings mentioned above, we conduct experiments on WOS with the bert-large-uncased (330M) encoder.           
            \textbf{Bold}: best results.
	}
\label{tab:model_scale}
\end{table}

\subsection{Performance Benefit in a Full-shot Setup}
We conduct experiments on HierVerb in a full-shot setting. Instead of carefully selecting hyperparameters, we directly use the parameter set from the few-shot settings.
For baseline models, we reproduce their experiments according to the settings in their original paper.
Although HierVerb is designed to be more favored for few-shot settings, the performance of full-shot setup is still quite competitive compared with HPT.
As shown in Table~\ref{tab:full_shot}, our overall micro-F1 score is only 0.10 lower than HPT (which requires to learn extra parameters of GNN), while achieving a macro-F1 score 0.13\% higher than HPT.
In fact, HierVerb outperforms BERT (Vanilla FT) and HiMatch by a significant margin. 

\begin{table}[!t]
   
    \centering
    \footnotesize
    \begin{tabular}{l|cc|cc}
    \toprule
    & \multicolumn{2}{|c|}{\textbf{WOS}}
    & \\
    \textbf{Methods}& Micro-F1 & Macro-F1 \\
    \midrule
        HierVerb &87.00  &\textbf{81.57}\\
        HPT  &\textbf{87.10}  &81.44\\
        HGCLR &87.08  &81.11 \\
        HiMatch &86.70  &81.06 \\
        BERT (Vanilla FT) &85.63  &79.07 \\
    \bottomrule
    \end{tabular}
     \caption{Full-shot experiments on WOS using bert-base-uncased.  
            \textbf{Bold}: best results.
     }
       \label{tab:full_shot}
\end{table}

\section{Conclusion}
In this paper, we define the few-shot settings on HTC tasks and a novel evaluation method based on the perspective of path correctness, which is valuable in practical applications.
We propose a novel approach to adapt flat prior knowledge in PLM to downstream hierarchical tasks.
The proposed HierVerb learns hierarchical-aware verbalizers through flat contrastive learning and constraint chain, which elegantly leverages the prior knowledge of PLMs for better few-shot learning.
We perform few-shot settings on HTC tasks and extensive experiments show that our method achieves state-of-the-art performances on 3 popular HTC datasets while guaranteeing excellent consistency performance.

\section*{Limitations}
Since the appearance of large pre-trained models such as
GPT-3 \citep{NEURIPS2020_1457c0d6}, there has been a wave of using large models without fine-tuning to do in-context learning directly to complete various NLP tasks, or to freeze the parameters of large models and then only optimize task-oriented parameters.
The proposed HierVerb is a lightweight method especially suitable for the case of insufficient labeled training data, but it is difficult to directly extend to a large-scale language model (i.e, >=175B) because large language models are hard to fine-tune in many situations.
In future work, we plan to study our method on a larger scale language model in which only parts of parameters specific to downstream HTC tasks need to be learned
and further, extend our model to the zero-shot learning scenario.


\section*{Ethics Statement}
All datasets for our research are publicly available and all experimental results are based on three different random seeds.
We obtain these experimental results using the experimental setup mentioned in this work.
For the sake of energy saving, we will not only open source the few-shot datasets under all random seeds and the code, but also release the checkpoints of our models from the experiments to reduce unnecessary carbon emissions.



\bibliography{anthology,custom}
\bibliographystyle{acl_natbib}

\clearpage

\appendix
\section{Implementation Details}
All our models are implemented with PyTorch \citep{paszke2019pytorch} framework, Huggingface transformers \citep{wolf-etal-2020-transformers}, and OpenPrompt toolkit \citep{ding-etal-2022-openprompt}.
Following previous work \citep{wang2022hpt}, we use \texttt{bert-base-uncased} from Transformers as our base architecture. 
The hidden size \textit{r} is 768, and the number of layers and heads are 12.
The batch size is 5. 
For WOS and DBPedia, the learning rate is $5e^{-5}$, besides we use a learning rate of $1e^{-4}$ to fasten the convergence of its hierarchical label words' embeddings and train the model for 20 epochs and apply the Adam Optimizer \citep{kingma2014adam} with a linearly decaying schedule with warmup steps at 0 and evaluate on the development set after every epoch.
Since the labels of RCV1 do not contain excessively rich natural text semantics, the training iteration on RCV1 is the same as HPT \citep{wang2022hpt} with 1000 epochs and we set early stopping to 10 and learning rate to 3$e^{-5}$ which is also used for the optimization of verbalizers.
For baseline models, we keep the hyperparameter settings from their original papers except for setting early stopping to 10 for a fair comparison. 
We list the details of the other hyperparameters in Table ~\ref{tab:implement_detail}.
\label{sec:imp_details}

\begin{table}[b]
\begin{center}
\resizebox{\linewidth}{!}{
\begin{tabular}{lll}
\hline
 Hyper-parameter &Dataset &Value \\ 
\hline
truncate length & All & 512\\
warmup steps & All & 0 \\
$\lambda$ 1 & All & 1 \\
$\lambda$ 2 & WOS\&DBPedia & 1e-2 \\
$\lambda$ 2 & RCV1-V2 & 1e-4 \\
$\alpha$ & All & 1 \\
$\beta$ & WOS\&DBPedia & 1\\
$\beta$ & RCV1-V2 & 1e-2\\
\hline

\end{tabular}
}
\end{center}
\caption{Hyper-parameter settings}
\label{tab:implement_detail}
\end{table}

\begin{figure}[t]
\centering
\includegraphics[width=0.5\textwidth]{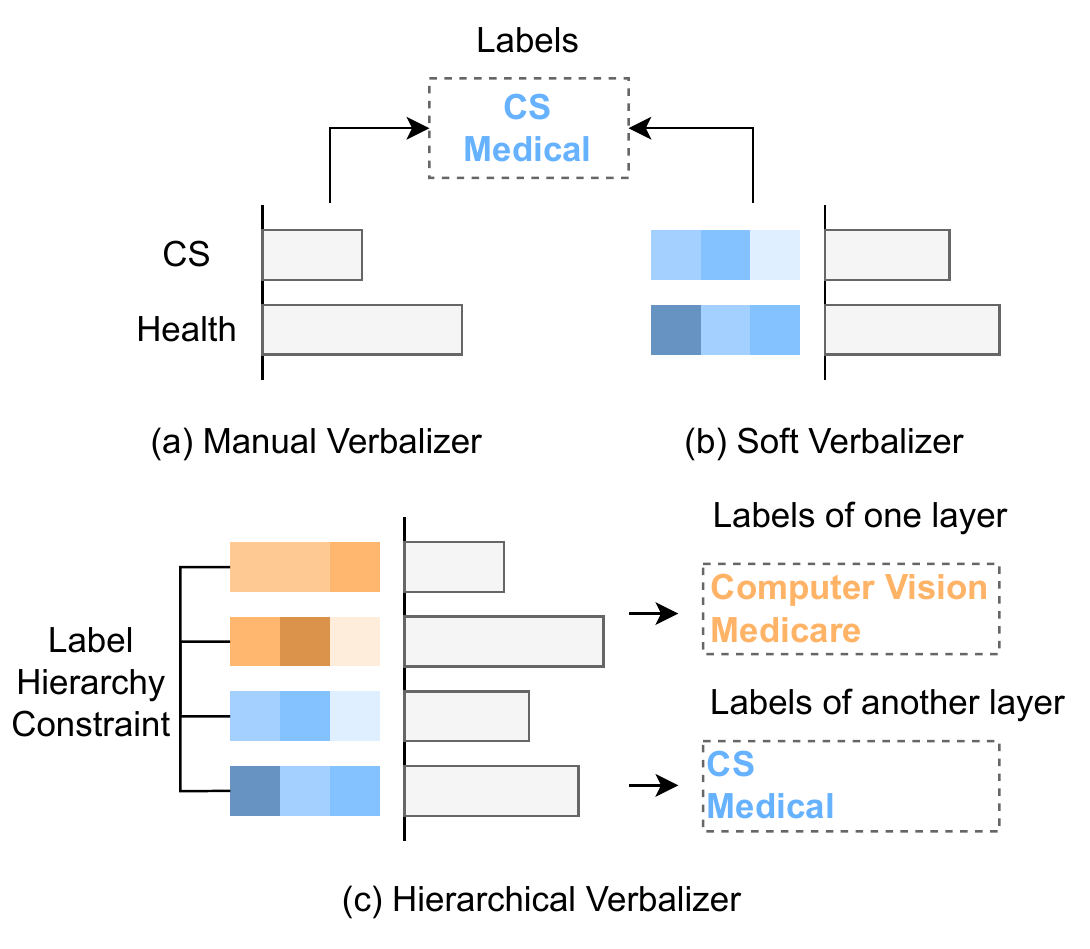}
\caption{Comparison between previous verbalizer design methods and our HierVerb: (a) Manual Verbalizer. (b) Soft Verbalizer utilizes learnable vectors for each label word. (c) Our HierVerb fuzes hierarchical constraints into the soft verbalizer, effective for hierarchical problems.}
\label{fig:hierverb2}
\end{figure}

\section{Path-based Evaluation Metric}
Specifically, in P-metric, we evaluate the confusion matrix of all label path ids instead of the original label ids. 
Besides, only if all $\{y_i\}$ labels involved in one path are predicted accurately, the corresponding path id is regarded as correct in the confusion matrix.
We count the total number of golden labels as $Count_{gold}$ and at the same time record the predicted labels that do not form a complete path with other predicted labels as invalid and count their total as $Count_{invalid}$.

\begin{figure*}[ht]
    \begin{minipage}[t]{0.24\linewidth}
        \centering
        \includegraphics[width=\textwidth]{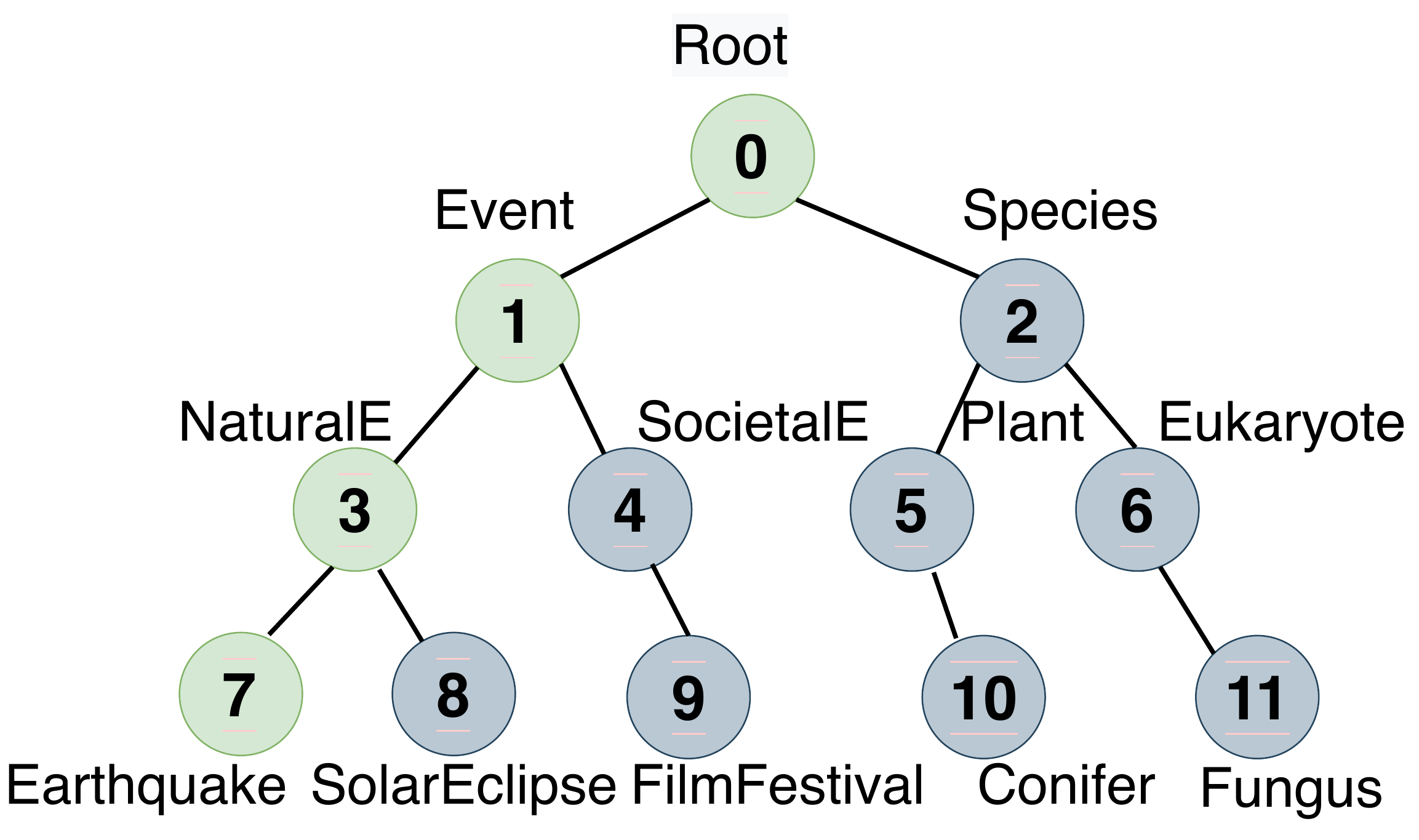}
        \centerline{(a)}
    \end{minipage}%
    \begin{minipage}[t]{0.24\linewidth}
        \centering
        \includegraphics[width=\textwidth]{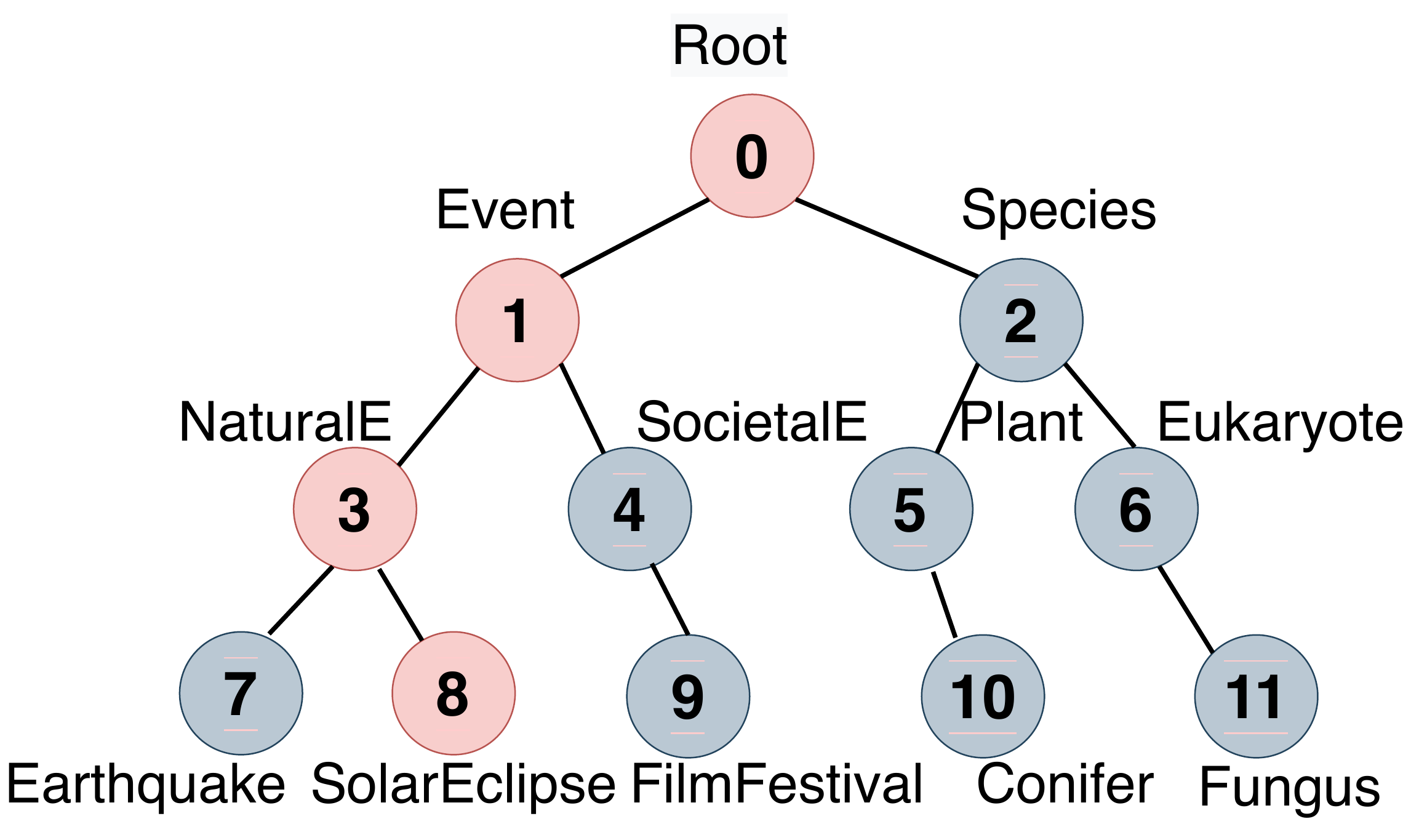}
        \centerline{(b)}
    \end{minipage}
    \begin{minipage}[t]{0.24\linewidth}
        \centering
        \includegraphics[width=\textwidth]{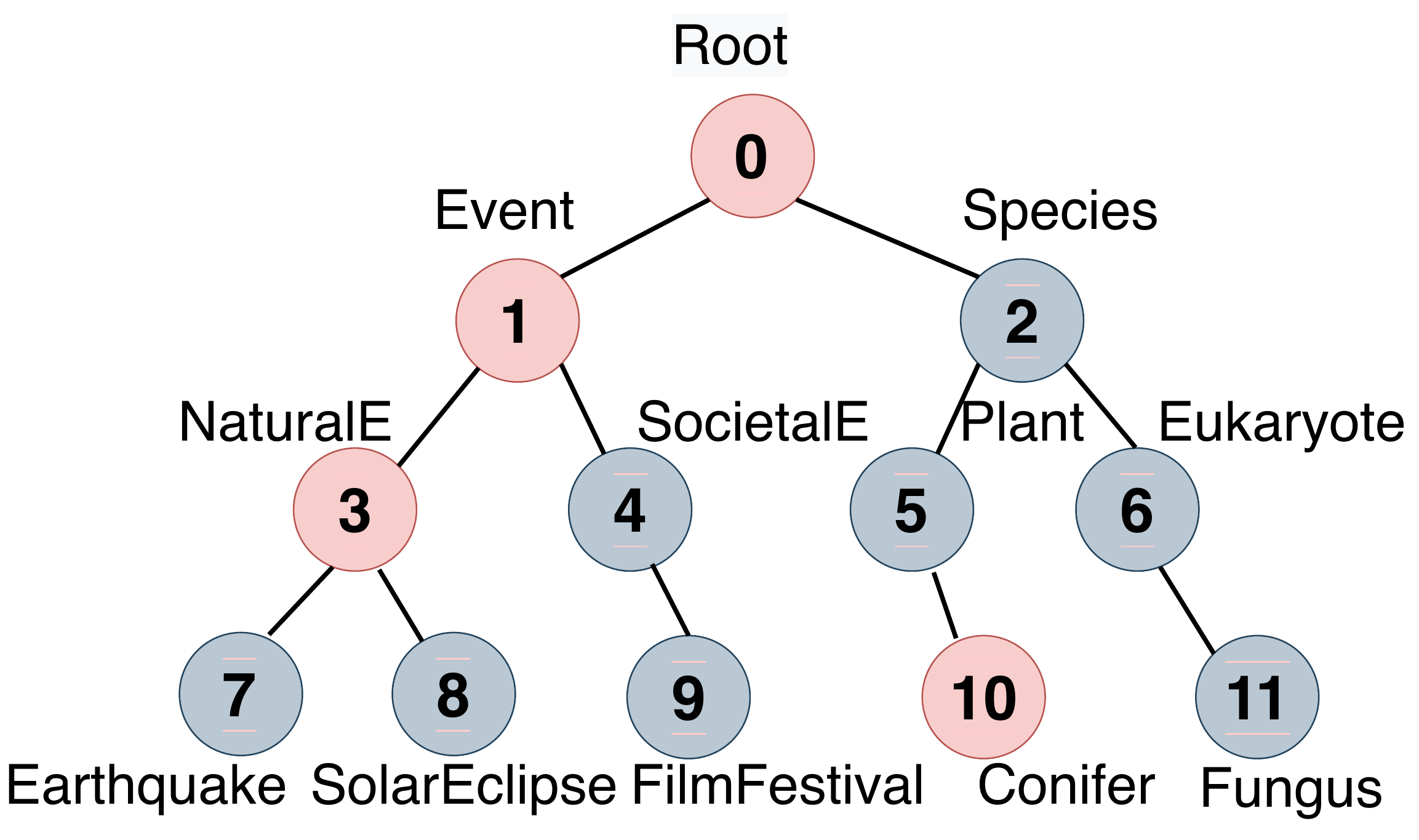}
        \centerline{(c)}
    \end{minipage}
    \begin{minipage}[t]{0.24\linewidth}
        \centering
        \includegraphics[width=\textwidth]{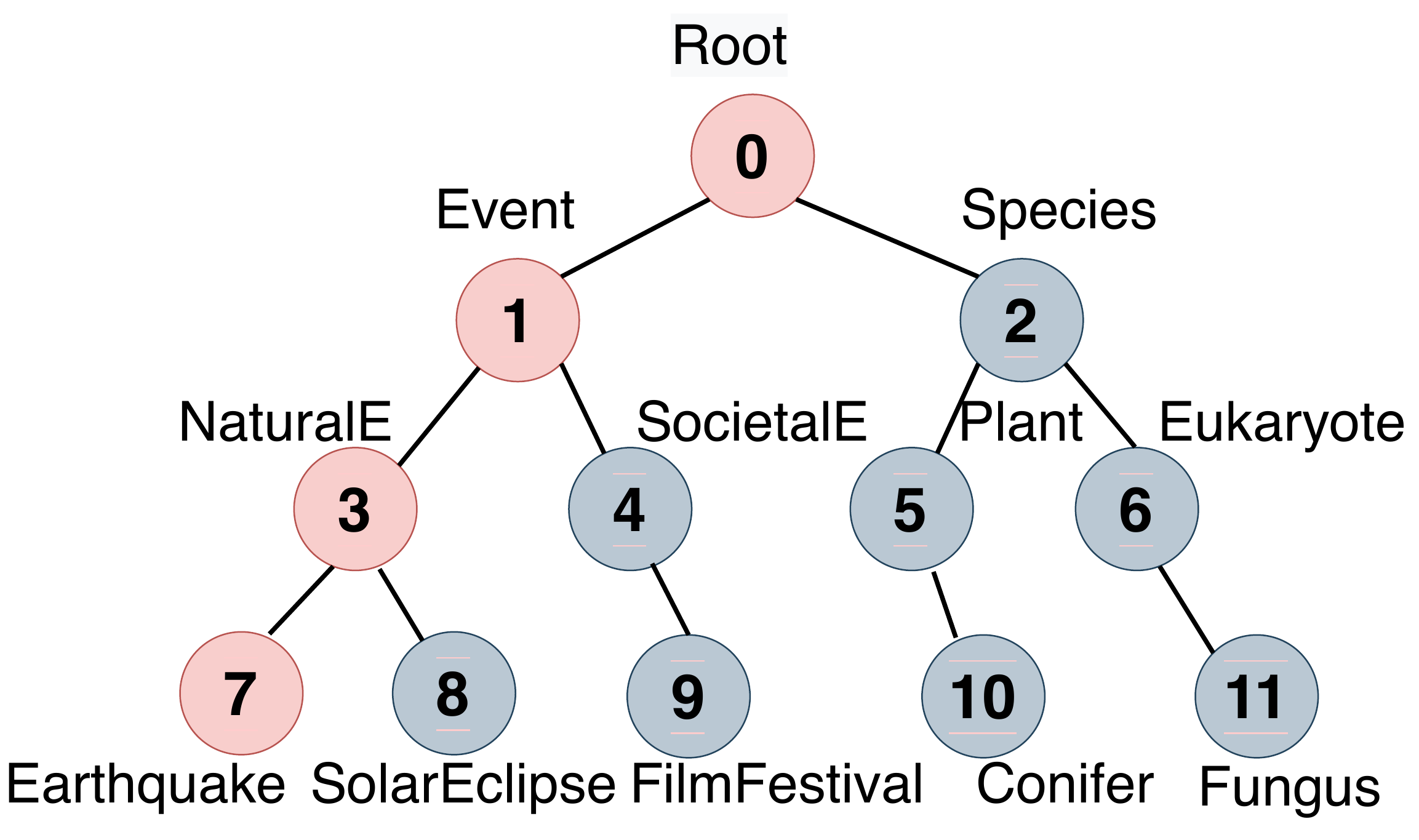}
        \centerline{(d)}
    \end{minipage}
    \caption{
    An example of taxonomic hierarchy for HTC.
    Assume we have an input sentence to predict, then we have (a) Golden labels of the input sentence, (b) and (c) Predicted labels lead to "label path inconsistency", and (d) Correctly predict all the labels in the path consisting of the node set \{1,3,7\}.
    We pick out these predicted labels in (c) that cannot form a complete path with any other predicted labels and record them as $\textit{Y}_{invalid}$=\{1,3,10\}, then we calculate the size of the set $Y_{invalid}$ and add it to $Count_{invalid}$.
    }
\label{fig:P_metric}
\end{figure*}

We define:
\begin{equation}
    \gamma = 1-2\times(\frac{1}{(1+e^{-a})} - 0.5) \\
\end{equation}
where a = $\frac{Count_{invalid}}{Count_{gold}}$ and multiply $\gamma$ with PMacro-F1 and PMicro-F1 obtained from the confusion matrix to get our final PMacro-F1 and PMicro-F1 so that we can penalize the evaluation score to get a fairer evaluation when the model smartly predicts a particularly large number of labels that do not form a complete path, considering that we are building confusion matrix based on the path.
Figure \ref{fig:P_metric} shows the inconsistency problem.
\label{sec:path_metric}

\begin{table*}[ht]
	\centering
	\footnotesize
	\setlength{\tabcolsep}{1.5mm}
	\begin{tabular}{clcccccccc}
		\toprule[1pt]
		\multicolumn{1}{c}{\multirow{2}{*}{\begin{tabular}[c]{@{}c@{}} \\ 
            K
        \end{tabular}}} & \multicolumn{1}{l}{\multirow{2}{*}{\begin{tabular}[c]{@{}c@{}} \\ 
            Method\end{tabular}}}
		& \multicolumn{4}{c}{\textbf{DBPedia}} & \multicolumn{4}{c}{\textbf{RCV1-V2}} \\ \cmidrule {3-10} 
		& & PMicro-F1 & PMacro-F1 & CMicro-F1 & CMacro-F1 & PMicro-F1 & PMacro-F1 & CMicro-F1 & CMacro-F1 \\ \midrule
  
            \multicolumn{1}{c}{\multirow{4}{*}{1}} 
		& \multicolumn{1}{|l}{Ours} &\textbf{83.56}  &\textbf{77.96} &\textbf{89.80} &\textbf{81.78} &-  &- &\textbf{39.41} &\textbf{5.16} \\
		& \multicolumn{1}{|l}{HPT} &61.08  &57.80 &82.84 &66.99 &- &- &21.92 &2.87 \\
		& \multicolumn{1}{|l}{HGCLR} &0.0  &0.0 &28.05 &0.24 &- &- &23.26 &1.04 \\
		& \multicolumn{1}{|l}{Vanilla FT} &0.0  &0.0  &28.08 &0.24 &- &- &19.37 &1.02\\
		\midrule
		
		\multicolumn{1}{c}{\multirow{4}{*}{2}} 
		& \multicolumn{1}{|l}{Ours} &\textbf{88.58}  &\textbf{86.35} &\textbf{93.61} &\textbf{88.96} &-  &- &\textbf{45.11} &\textbf{12.32} \\
		& \multicolumn{1}{|l}{HPT} &82.36  &81.41 &92.31 &86.43 &- &- &38.24 &7.00\\
		& \multicolumn{1}{|l}{HGCLR} &54.55  &3.72 &67.70 &26.41  &- &- &24.24 &0.89\\
		& \multicolumn{1}{|l}{Vanilla FT} &53.83  &3.71 &67.72 &26.89 &- &- &23.60 &0.81 \\
		\midrule
		
		\multicolumn{1}{c}{\multirow{4}{*}{4}} 
		& \multicolumn{1}{|l}{Ours} &\textbf{91.90}  &\textbf{91.38} &\textbf{95.74}  &\textbf{92.87}  &-  &- &\textbf{54.67}  &\textbf{23.80}\\
		& \multicolumn{1}{|l}{HPT} &87.61  &87.04 &94.50 &90.42 &- &- &50.68 &20.54\\
		& \multicolumn{1}{|l}{HGCLR} &55.34  &3.76  &67.54  &28.60 &- &- &44.74 &9.02\\
		& \multicolumn{1}{|l}{Vanilla FT} &55.15  &3.74  &67.44 &28.32 &- &- &22.42 &0.63\\

		\bottomrule[1pt]
	\end{tabular}
	\caption
	{
    	\label{Result} 
    	Consistency experiments on the DBPedia and RCV1-V2 datasets using two path-constraint metrics.
            PMicro-F1 and PMacro-F1 are our proposed path-based consistency evaluation P-metric.
            Since the label distribution of the original test set of RCV1-V2 is not mandatory-leaf in $\mathcal{H}$ while WOS and DBPedia are, we use only C-metric on RCV1-V2 to evaluate the consistency performance.
            We report the mean F1 scores (\%) over 3 random seeds.
            \textbf{Bold}: best results.
            All experiments use their respective metrics as a signal for early stopping.
	}
\label{tab:consistency_complete}
\end{table*}

\begin{figure}[t]
    \begin{minipage}[t]{0.50\linewidth}
        \centering
        \includegraphics[width=\textwidth]{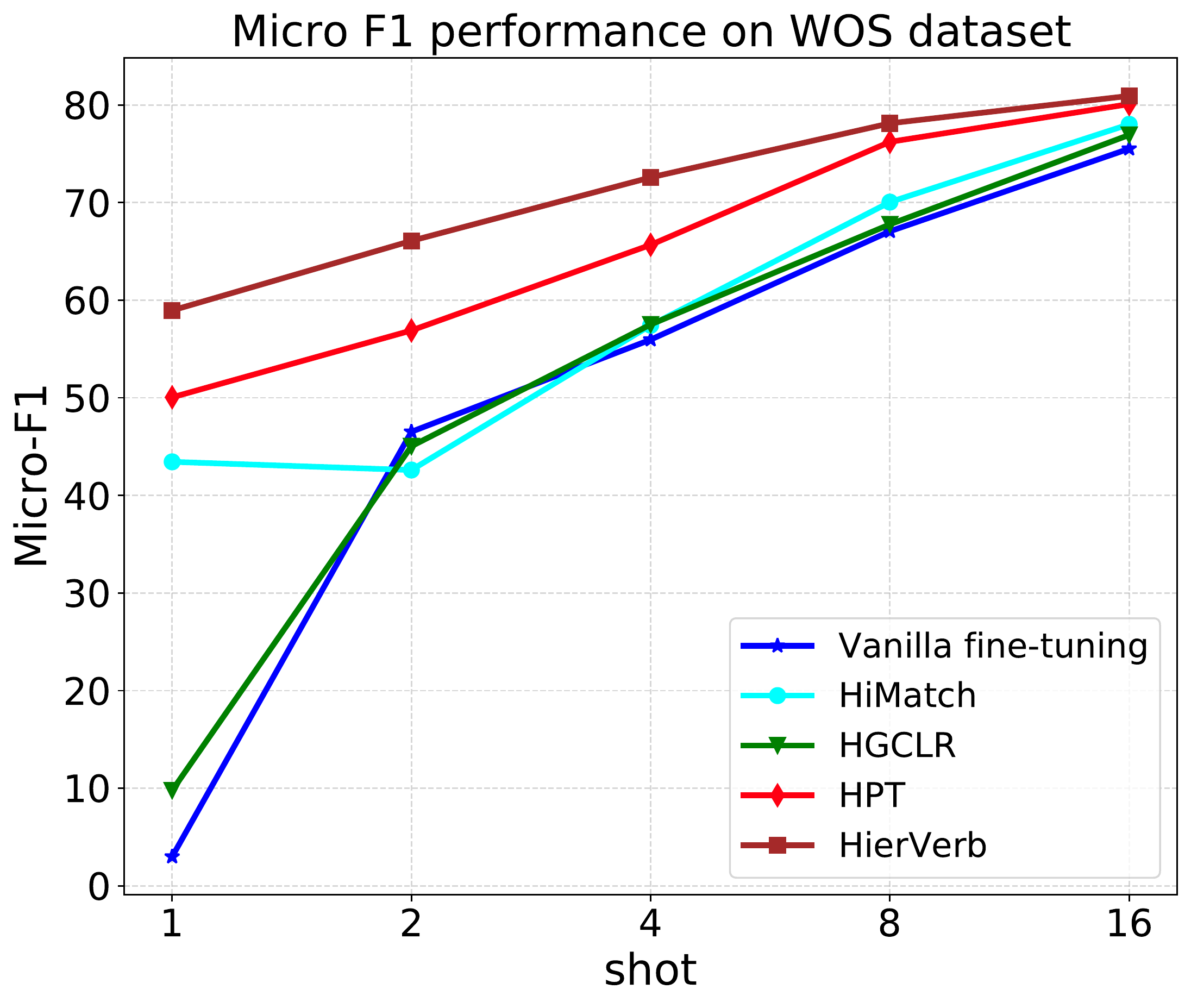}
        \centerline{(a)}
    \end{minipage}%
    \begin{minipage}[t]{0.50\linewidth}
        \centering
        \includegraphics[width=\textwidth]{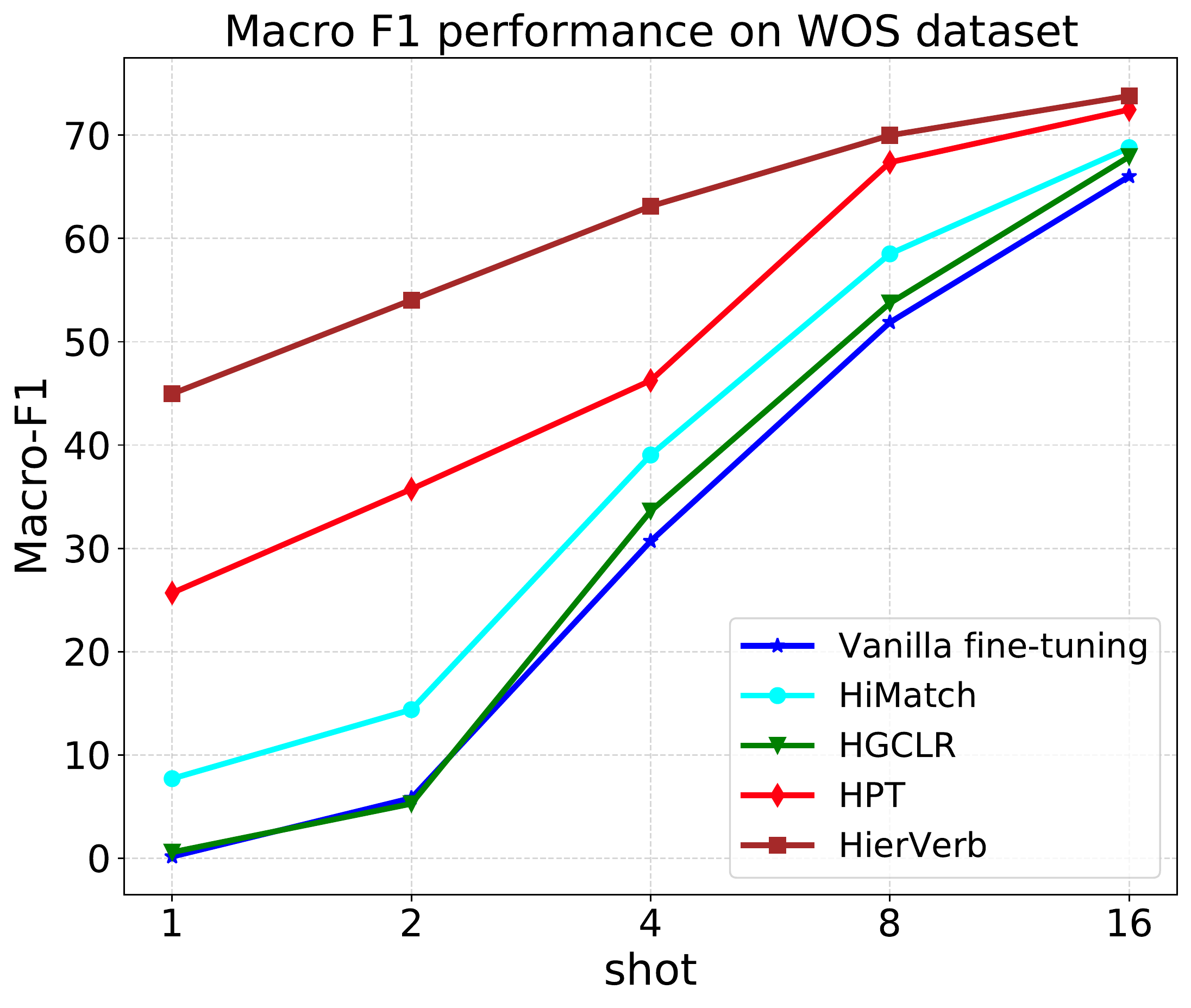}
        \centerline{(b)}
    \end{minipage}
    \caption{Performance gap on WOS dataset}
\label{fig:comparison_wos}
\end{figure}
\begin{figure}[!t]
    \begin{minipage}[t]{0.50\linewidth}
        \centering
        \includegraphics[width=\textwidth]{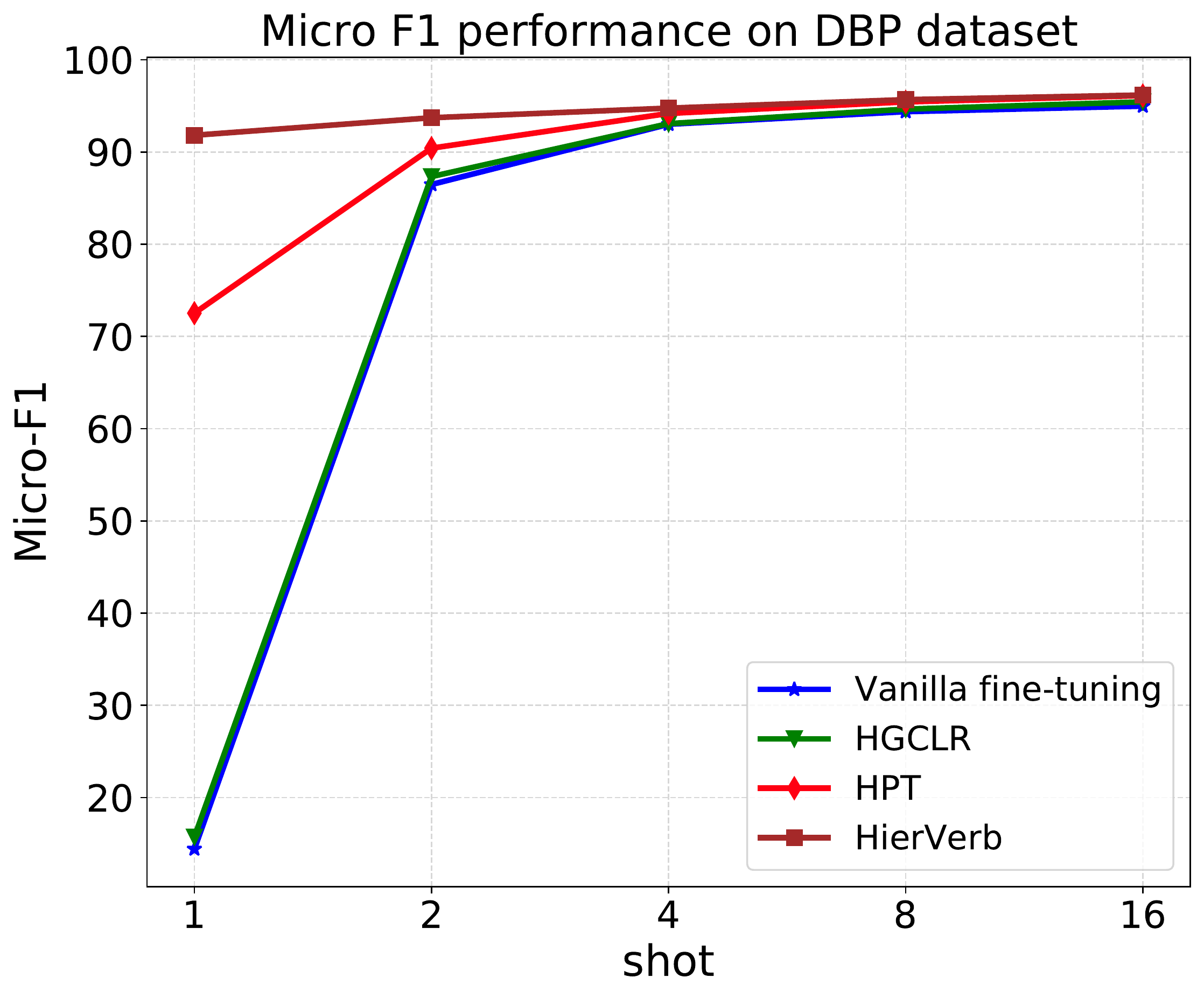}
        \centerline{(a)}
    \end{minipage}%
    \begin{minipage}[t]{0.50\linewidth}
        \centering
        \includegraphics[width=\textwidth]{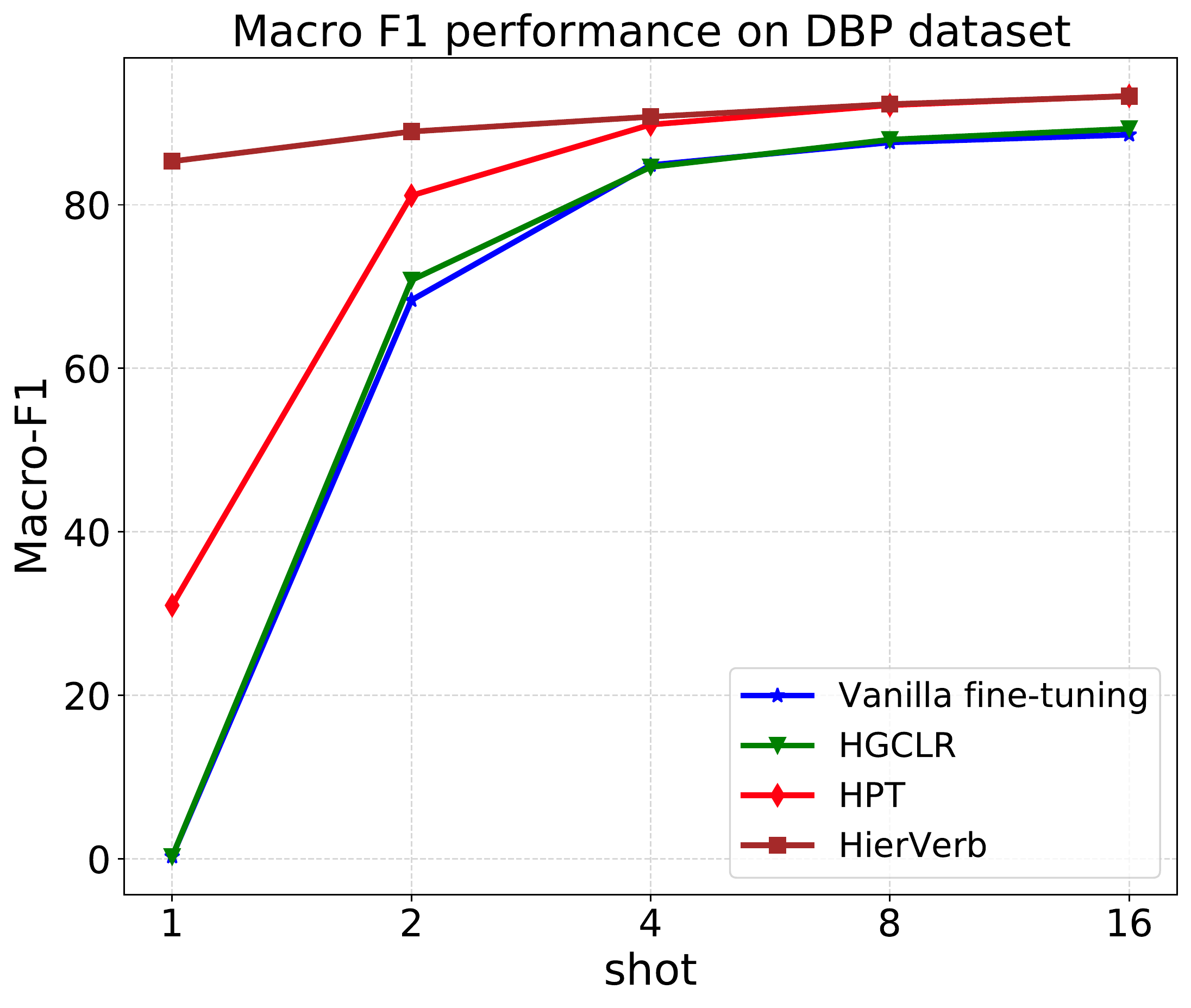}
        \centerline{(b)}
    \end{minipage}
    \caption{Performance gap on DBPedia dataset}
\label{fig:comparison_dbp}
\end{figure}
\begin{figure}[t]
    \begin{minipage}[t]{0.50\linewidth}
        \centering
        \includegraphics[width=\textwidth]{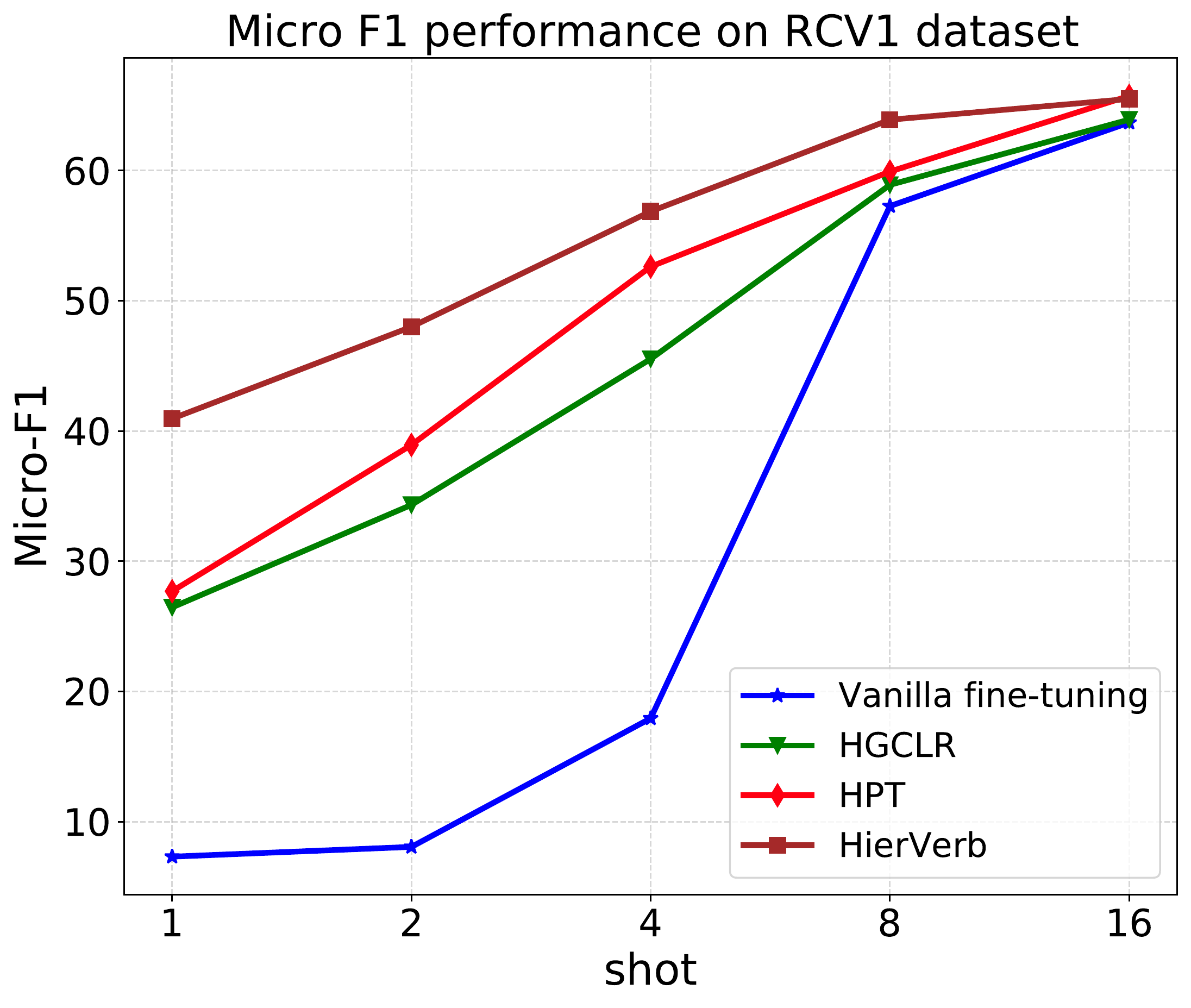}
        \centerline{(a)}
    \end{minipage}%
    \begin{minipage}[t]{0.50\linewidth}
        \centering
        \includegraphics[width=\textwidth]{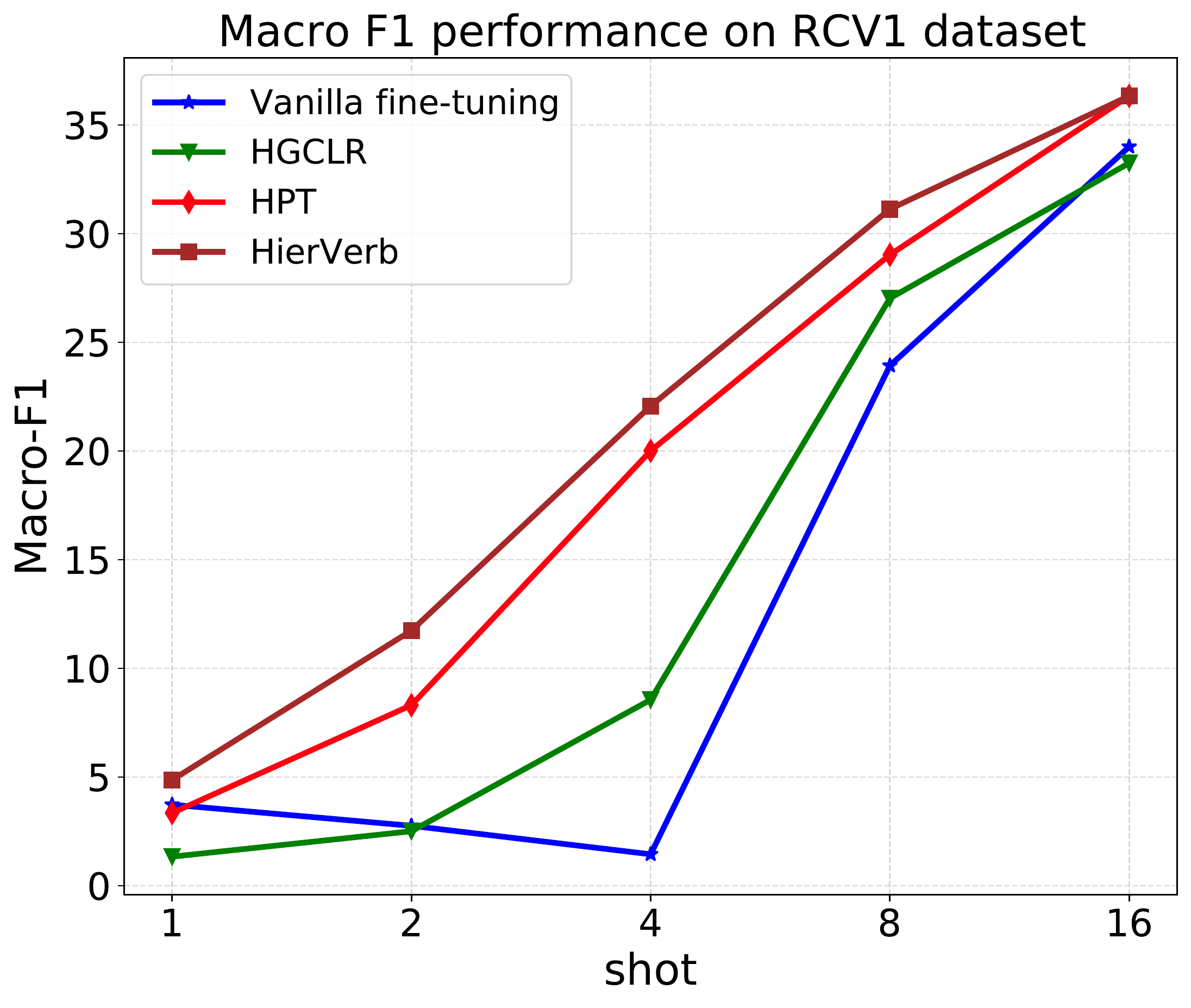}
        \centerline{(b)}
    \end{minipage}
    \caption{Performance gap on RCV1-V2 dataset}
\label{fig:comparison_rcv1}
\end{figure}

\section{Performance Gap between Different Models}
The performance gap on three datasets between different models is clearly shown in Figure \ref{fig:comparison_wos}-\ref{fig:comparison_rcv1}.
The gap keeps growing as the shots become fewer.
It can be clearly seen that both HierVerb's Micro-F1 and Macro-F1 change very slightly from 1 to 16 shots on DBPedia while other models are particularly dependent on the increase of labeled training samples.
\label{sec:performance_gap}

\section{Complete Consistency Experiments}
We further conduct consistency experiments on two other datasets.
The results are shown in Table~\ref{tab:consistency_complete}.
In all experiments, HGCLR and Vanilla FT consistently perform poorly on both P-Metric and C-Metric, while HierVerb and HPT achieved relatively high results, indicating that the prompt-based method can better use the prior knowledge in the pre-trained model to elicit
potential semantic associations between natural language texts of all labels belonging to the same path.
\label{sec:consistency_experiments}

\end{document}